\documentclass{article}

\usepackage{microtype}
\usepackage{graphicx}
\usepackage{subcaption}
\usepackage[utf8]{inputenc} 
\usepackage[T1]{fontenc}    

\usepackage{url}
\usepackage{amsfonts}
\usepackage{nicefrac}       
\usepackage{microtype}      
\usepackage{lipsum}		
\usepackage{graphicx}
\usepackage{tikz}
\usepackage{tikz-dependency}
\usepackage{tikz-qtree}
\usepackage[round]{natbib}
\usepackage{doi}
\usepackage{colortbl,booktabs}
\usepackage{xcolor}
\usepackage{wrapfig}
\usepackage{environ}
\usepackage{multicol}
\usepackage{multirow}
\usepackage{wrapfig}
\usepackage{makecell}
\usepackage{diagbox}
\usepackage{pifont}
\usepackage{pgffor}
\usepackage[english]{babel}
\usepackage{xspace}
\usepackage{forest}
\usepackage{amsmath}
\usepackage{verbatim}

\usepackage{pgfplots}
\usepgfplotslibrary{dateplot}
\usepackage{xspace}
\usepackage{array}
\usepackage{colortbl}
\usepackage{xcolor}

\usepackage{tcolorbox}
\tcbuselibrary{skins}
\newtcolorbox{promptbox}{
  enhanced,
  colback=gray!5,
  colframe=blue!75!black,
  boxrule=1pt,
  arc=4pt,
  left=6pt,
  right=6pt,
  top=6pt,
  bottom=6pt,
  fontupper=\ttfamily,
  title=Prompt
}

\definecolor{lightblue}{rgb}{0.9, 0.95, 1.0}
\definecolor{almond}{rgb}{0.94, 0.87, 0.8}
\definecolor{amber}{rgb}{1.0, 0.75, 0.0}
\definecolor{alizarin}{rgb}{0.82, 0.1, 0.26}
\definecolor{antiquewhite}{rgb}{0.98, 0.92, 0.84}
\definecolor{codegray}{RGB}{128, 128, 128}
\newcolumntype{g}{>{\columncolor{codegray!25}}r}

\usepackage{hyperref}

\DeclareMathOperator*{\argmax}{arg\,max}

\usepackage[accepted]{icml2025}

\usepackage{amsmath}
\usepackage{amssymb}
\usepackage{mathtools}
\usepackage{amsthm}
\usepackage{graphicx}
\usepackage{algorithmic}
\usepackage{algorithm}
\usepackage[capitalize,noabbrev]{cleveref}

\theoremstyle{plain}

\theoremstyle{definition}

\theoremstyle{remark}

\usepackage[textsize=tiny]{todonotes}

\definecolor{Brown}{rgb}{0.65, 0.16, 0.16}
\definecolor{RoyalBlue}{RGB}{65, 105, 225}
\definecolor{mycite}{cmyk}{0.55,1,0,0.15}
\definecolor{lightgray}{rgb}{0.9, 0.9, 0.9}

\hypersetup{
  colorlinks=true,
  citecolor=mycite,
  urlcolor=Brown,
  linkcolor=Brown,
}

\def\SC{SynContextQA\xspace}
\def\SG{SimGSM8K\xspace}
\def\LIMA{LIMA\xspace}

\icmltitlerunning{Unnatural Languages Are Not Bugs but Features for LLMs}

\begin{document}

\twocolumn[
	\icmltitle{Unnatural Languages Are Not Bugs but Features for LLMs}



	\icmlsetsymbol{equal}{*}

	\begin{icmlauthorlist}
		\icmlauthor{Keyu Duan}{equal,NUS}
		\icmlauthor{Yiran Zhao}{equal,NUS}
		\icmlauthor{Zhili Feng}{CMU}
		\icmlauthor{Jinjie Ni}{NUS}
		\icmlauthor{Tianyu Pang}{Sea}
		\icmlauthor{Qian Liu}{Sea}
		\icmlauthor{Tianle Cai}{prin}
		\icmlauthor{Longxu Dou}{Sea}
		\icmlauthor{Kenji Kawaguchi}{NUS}
            \icmlauthor{Anirudh Goyal}{Mila}
		\icmlauthor{J. Zico Kolter}{CMU}
		\icmlauthor{Michael Qizhe Shieh}{NUS}
	\end{icmlauthorlist}

	\icmlaffiliation{NUS}{National University of Singapore}
	\icmlaffiliation{CMU}{Carnegie Mellon University}
	\icmlaffiliation{Sea}{Sea AI Lab}
        \icmlaffiliation{prin}{Princeton University}
        \icmlaffiliation{Mila}{Mila, University of Montreal}

        \icmlcorrespondingauthor{Keyu Duan}{k.duan@u.nus.edu}
	\icmlcorrespondingauthor{Yiran Zhao}{zhaoyiran@u.nus.edu}
        \icmlcorrespondingauthor{Michael Qizhe Shieh}{michaelshieh@comp.nus.edu.sg}

	\icmlkeywords{Machine Learning, ICML}

	\vskip 0.3in
]



\printAffiliationsAndNotice{\icmlEqualContribution} 

\begin{abstract}
	Large Language Models (LLMs) have been observed to process non-human-readable text sequences, such as jailbreak prompts, often viewed as a bug for
	aligned LLMs. In this work, we present a systematic investigation challenging this
	perception, demonstrating that unnatural languages - strings that appear
	incomprehensible to humans but maintain semantic meanings for LLMs - contain latent features usable by models. Notably, unnatural languages possess latent features that can be generalized
across different models and tasks during inference. Furthermore, 
	models fine-tuned on unnatural versions of instruction datasets perform on-par
	with those trained on natural language, achieving \(49.71\) win rates in
	Length-controlled AlpacaEval 2.0 in average across various base models.
	In addition, through comprehensive analysis,  we demonstrate that LLMs process unnatural languages by filtering noise and inferring contextual meaning from filtered words. Our code is publicly available at \url{https://github.com/John-AI-Lab/Unnatural_Language}.
\end{abstract}

\section{Introduction}
Large Language Models (LLMs)~\citep{john2023chatgpt,touvron2023llama,dubey2024llama,anthropic2024claude}
have shown remarkable capabilities in understanding and generating human-readable text,
achieving impressive performance across tasks, spanning from question
answering~\citep{bisk2020piqa,ni2024mixeval} and mathematical reasoning~\cite{cobbe2021training,hendrycks2021measuring,gao2024omni} to open-ended
dialogue~\citep{alpaca_eval}. Such abilities are largely attributed to targeted alignment
training~\citep{wei2021finetuned,ouyang2022training}, which post-train models to better
follow instructions and adhere to preferred behaviors.

Despite being specifically tuned, non human-readable data sometimes can
unexpectedly influence model behavior. In computer vision,
\citet{ilyas2019adversarial,nguyen2015deep} find that seemingly unrecognizable
images could be leveraged to train reasonable good image classification models.
This phenomenon extends to natural language processing, where
\citet{zou2023universal} demonstrate that LLMs could also be prompted with an
unreadable suffix to generate objectionable outputs, even though LLMs are
well-trained for not doing so. Besides, \citet{pfau2024let} discovers that by
appending non human-readable filler tokens to the input, LLMs could solve
algorithmic tasks more accurately. However, regarding LLMs' surprising behaviors
in response to non human-readable inputs, there lacks systematic studies
exploring the properties and applications of such non human-readable strings and
interpreting the underlying mechanisms in LLMs. This raises a fundamental
question: \textit{whether these non human-readable strings are truly devoid of
	meaning or contain latent features usable by models?}

\begin{table}[t]
	\caption{Concrete examples of natural contexts' unnatural version in SycContextQA and SimGSM8K. }
	\centering
	\scalebox{0.9}{
		\begin{tabular}{p{3cm}|p{5cm}}
			\toprule
			 \textbf{Natural Context} & \textbf{Unnatural Version}  \\\midrule
			 The stock price of GoldMine Inc. increased by 20\% last week.             & (alt+eqn=\{\textbackslash\textbackslash >; \{\};The\textbackslash\textbackslash ,\textbackslash\textbackslash stock baaelkrie@nuier priceungeureau got sich last '\#GM;;heidisation Inc. weekestig \%\}20\% durch'),png encrypt render \textbackslash{}\textquotedbl{}OK GoldMine.\textquotedbl{},preventDefault  \\\midrule
			 Carly collected 7 starfish with 5 arms each and one seastar with 14 arms. & |Each        and : algebra dinner! absolutely         7 do): shortly . seastar collectedthe `{}' kW)\$, one !5 ! 14`{} starfish with sic\}\}\_\{\textbackslash{}label Carly\} arms. Onehorailey constructed WriteStatus(\$\$\textbackslash{}Toggle Zwezeichnung OK                                                                    \\
			\bottomrule
		\end{tabular}}
	\label{table:dataset}
\vspace{-0.3cm}
\end{table}

\begin{figure*}[t]
	\centering
	\includegraphics[width=0.98\textwidth]{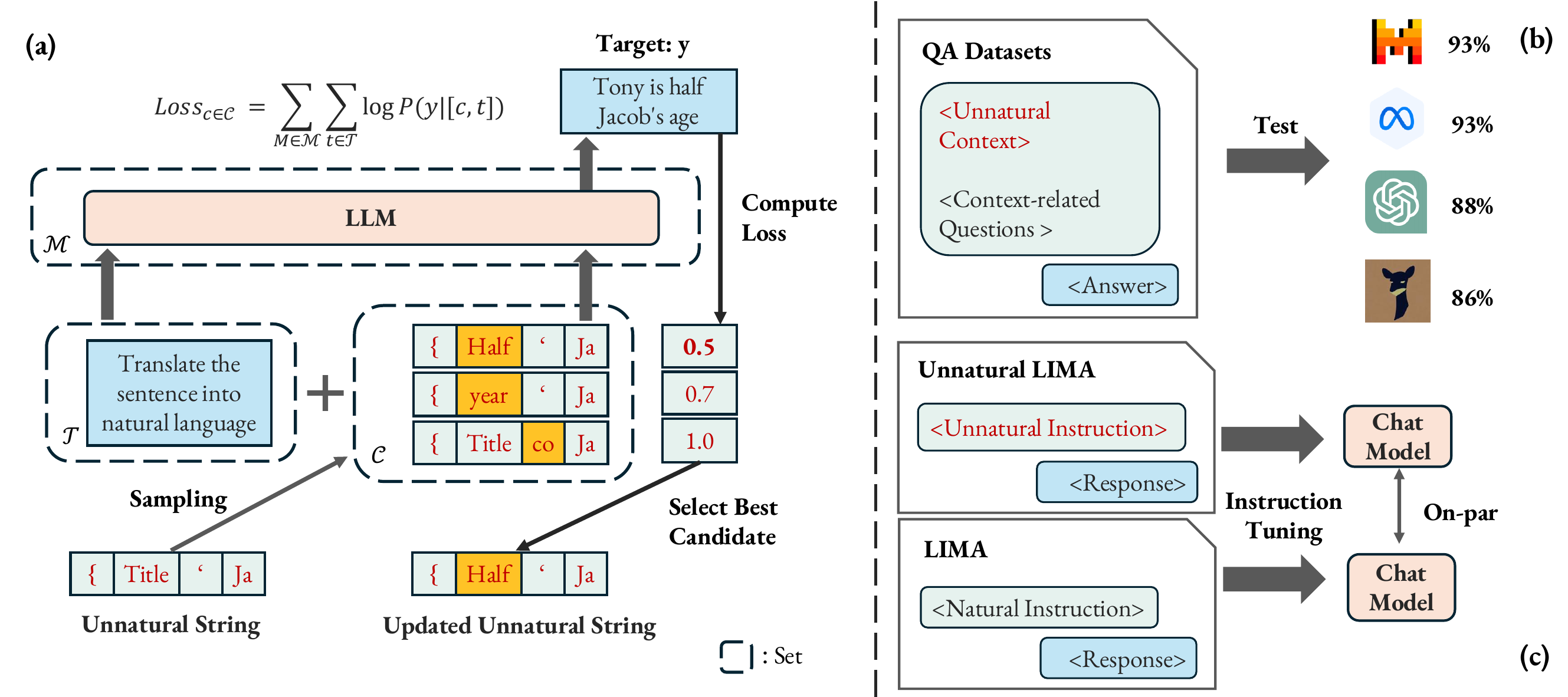}
    \vspace{-0.2cm}
	\caption{\textit{(a)}: Unnatural languages searching method. \textit{(b)}: We construct question-answering tasks using unnatural contexts and discover that unnatural languages can be directly transferred across a broader range of tasks and understood by diverse LLMs. (c): LLMs fine-tuned on unnatural versions of instruction datasets perform on-par with those trained on natural language.}
	\label{fig:framework}
    \vspace{-0.2cm}
\end{figure*}

To answer the above research questions, we study a phenomenon named \textit{unnatural languages} - strings that deviate
from natural language syntax and appear extremely noisy to human readers, yet remain understandable to LLMs.
Specifically, as illustrated in Figure~\ref{fig:framework}(a), we propose an approach to search for a semantically
equivalent but syntactically unnatural version of the natural string, where semantic equivalence is established through
their ability to be translated back to natural form via models performing translation inference.
For the searching process, we employ a gradient-based stochastic sampling procedure to obtain a set of candidate
unnatural strings; evaluate their probability of being translated back to the corresponding natural versions across
multiple models; and select the unnatural candidate that yields the highest probability. We repeat the process until
either convergence occurs or a maximum number of iterations is reached, with the final converged string representing the
semantically equivalent unnatural string.

We first explore whether these unnatural languages possess latent features that
can be generalized across different models and tasks during inference.  To
investigate this, as shown in Figure \ref{fig:framework}(b), we construct
several context-based question-answering datasets, where the context is
unnatural obtained by the searching approach while natural questions
related to the context are provided. Specifically, to prevent models from
relying on common-sense memory when answering questions without context, we
develop SynContextQA, a synthetic dataset generated by another LLM, containing
contexts about non-existent entities paired with corresponding questions. We
then transformed natural contexts into unnatural versions using the unnatural
language searching method while preserving the original questions. Additionally,
to ensure models do not simply extract keywords from unnatural contexts in
SynContextQA, we create SimGSM8K, a dataset of simple questions derived from
GSM8K~\citep{cobbe2021training}. We chose simple questions to minimize the
impact of reasoning ability on our results and focus primarily on unnatural
language comprehension. As with SynContextQA, we transformed these contexts into
unnatural versions. Table \ref{table:dataset} shows concrete examples of two datasets. With these datasets, we test a large variety of LLMs, including open-source models as well as commercial models. The results show that compared to natural context,
all models can recover $82.0\%$ of the original accuracy on our constructed
SynContextQA dataset and $61.6\%$ on SimGSM8K, demonstrating that unnatural
languages contain latent features that enable comprehension across different
scenarios.


Moreover, we explore whether these unnatural languages possess transferable latent features that can be effectively
utilized in instruction tuning to improve models' instruction-following capabilities. Specifically, as shown in Figure \ref{fig:framework}(c), we employ a
high-quality but small size instruction tuning dataset LIMA~\citep{zhou2023lima}, and we replace the original
instructions with our equivalent unnatural versions searched using our proposed approach. We show that the models
fine-tuned on it and the original have on-par performance on prestigious benchmarks, including Length-controlled (LC)
AlpacaEval 2.0~\citep{alpaca_eval} and MixEval~\citep{ni2024mixeval}. Particularly, on LC AlpacaEval 2.0, the three
models --- Llama-3-8B~\citep{dubey2024llama}, Gemma-2-9B~\cite{team2024gemma}, and Llama-3-70B~\citep{dubey2024llama}
--- tuned on the unnatural LIMA achieves an winrate of \(49.78\%\), \(47.13\%\), and \(52.22\%\) against the
corresponding models tuned on the natural LIMA, respectively.

These findings strongly demonstrate our key findings: unnatural languages are not bugs but features for LLMs. In
addition, we attempt to understand the mechanisms by which LLMs process such unnatural languages. We demonstrate that
LLMs process unnatural languages by effectively filtering out irrelevant tokens.
Furthermore, LLMs combine relevant tokens from unnatural languages and infer contextual meaning in response to natural version questions.

\section{Unnatural Languages Searching Method}~\label{sec:method}

\vspace{-0.3cm}

In this section, we introduce our approach for searching the unnatural version
of any given natural string.

\noindent\textbf{Problem description.} \;
We denote a natural string as $S$ and its equivalent unnatural version as ${S}'$. The equivalence between them is formally defined as ${S} \equiv \mathcal{LLM}_{M}({S}'|t)$, where $t$ represents a reconstruction task—such as translating the unnatural sentence into natural language—and $\mathcal{LLM}_M(S'|t)$ denotes the output of model $M$ given input $S'$ under task prompt $t$. Furthermore, we define the log-probability of model $M$ generating natural string $S$ when given unnatural string $x$ under task prompt $t$ as $\log P_M(S|x,t)$. Therefore, the unnatural string searching problem can be formulated as
\begin{equation}
	S' \coloneqq \argmax_{x\in \mathcal{X}} \log P_M(S| x, t),
	\label{eq:searching_obj_sing}
\end{equation}
where $\mathcal{X}$ represents the unnatural languages space, encompassing all possible strings of arbitrary length. However, searching the entire unnatural space is computationally infeasible. For simplicity and without loss of generality, we constrain $x$ to maintain a fixed length at the token level, i.e., $x\in \mathcal{X}_M$, where $ \mathcal{X}_M\triangleq\{x\big||\text{tokenize}_M(x)| = n, x\in\mathcal{X}\}$ and $n$ is a predefined constant length.

Furthermore, to enhance the generalizability of the obtained unnatural languages, we employ multiple models, denoted as $\mathcal{M}=\{M_1, M_2, \dots, M_k\}$, to collaboratively search for unnatural strings. Additionally, we introduce a set of tasks $\mathcal{T}=\{t_1, t_2, \dots, t_m\}$ as a collaborative optimization objective. Therefore, our goal of searching the equivalent unnatural string $S'$ is formulated as solving the following optimization problem:
\begin{equation}
	S' \coloneqq \argmax_{x \in \bigcup_{M\in \mathcal{M}}\mathcal{X}_M } \sum_{M \in \mathcal{M}} \sum_{t \in \mathcal{T}} \log P_M(S | x, t).
	\label{eq:searching_obj_multi}
\end{equation}

\noindent\textbf{Algorithm description.} \;
The optimization problem defined in Equation \ref{eq:searching_obj_multi} is a discrete optimization problem as $\mathcal{X}_M$ is a discrete space of size $|\mathcal{V}_M|^n$, where $\mathcal{V}_M$ denotes the vocabulary set of the model $M$. Due to the discrete nature of the problem, gradient-based optimization methods cannot be directly applied. Furthermore, the search space is too vast for exhaustive exploration. Therefore, we propose a sample-and-selection algorithm inspired by the optimization approaches of \citet{shin2020autoprompt,zou2023universal}. Specifically, in each optimization iteration, the unnatural string $x$ is first tokenized by model $M$ into $\mathbf{x}_{1:n}$. For each position, we identify the top-k most influential tokens $\mathbf{X}_{1:n}$ based on the gradient of the optimization objective of Equation \ref{eq:searching_obj_multi}, i.e., \begin{equation}
	\text{Top-k}\big(\nabla_{\mathbf{x}_{1:n}}\sum_{t\in\mathcal{T}}\log P_M(S | \mathbf{x}_{1:n}, t)\big).
\end{equation}
We then generate $B$ candidates $\{\tilde{\mathbf{x}}_{1:n}^{(1)}, \tilde{\mathbf{x}}_{1:n}^{(2)}, \cdots, \tilde{\mathbf{x}}_{1:n}^{(B)}\}$, where each candidate differs from $\mathbf{x}_{1:n}$ by exactly one token, randomly sampled from $\mathbf{X}_{1:n}$. These candidates are then decoded back to strings $\{\tilde{x}^{(1)}, \tilde{x}^{(2)}, \cdots, \tilde{x}^{(B)}\}$ by model $M$ for subsequent cross-model unification optimization. This candidate generation process is applied across all models in $\mathcal{M}$, yielding $B|\mathcal{M}|$ total candidates. The candidate with the optimal loss is selected for the next iteration. This process continues until convergence or until reaching a pre-defined number of iterations.





\noindent\textbf{Implementation details.} \;
In practice, tokenized sequences of $x$ have varying lengths across optimization
iterations due to different models employing distinct tokenizers, with no direct
one-to-one mapping between tokens and words. To maintain generalizability and in line with \citet{zou2023universal, zhao2024accelerating}, we initialize $x$ through a combination of
shuffling words in $S$ and randomly inserting several special characters ``!''. Algorithm \ref{alg:searching_ul} in Appendix \ref{sec:appen_alg} provides a
detailed illustration of the searching algorithm. In addition, we conduct verification experiments to demonstrate that our searched unnatural strings can be accurately translated back to their natural versions. Details are further illustrated in Appendix~\ref{sec:verification}.

\begin{table*}[t]
	\centering
	\caption{Performance comparison for different contexts across different models on SynContextQA and SimGSM8K datasets.
		All answers were generated under zero-shot setting without sampling. ``Direct'' refers to models used for unnatural languages searching, while ``Transfer'' indicates the implementation of searched unnatural languages.}
	\scalebox{0.95}{
		\begin{tabular}{ll|>{\columncolor{lightgray}}cc>{\columncolor{lightblue}}c|>{\columncolor{lightgray}}cc>{\columncolor{lightblue}}c}
			\toprule
			                                                & \multirow{2}{*}{\textbf{\normalsize{Model}}} & \multicolumn{3}{c}{\textbf{SynContextQA}} \vline & \multicolumn{3}{c}{\textbf{SimGSM8K}}                                     \\\cmidrule(lr){3-8}
			                                                &                                              & Natural                                          & Shuf-InJ                              & Unnatural & Natural
			                                                & Shuf-InJ                                     & Unnatural                                                                                                                    \\
			\midrule
			\multirow{3}{*}{\textbf{\normalsize{Direct}}}   & Mistral-7B-Instruct-v0.1                          & 0.89                                             & 0.55                                  & 0.93      & 0.85    & 0.20 & 0.42 \\
			                                                & Vicuna-7B-v1.5                                    & 0.96                                             & 0.40                                  & 0.86      & 0.63    & 0.12 & 0.20 \\
			\cmidrule(lr){2-8}
			                                                & \textit{Average}
			                                                & 0.93                                         & 0.48                                             & \textbf{0.90}                                  & 0.74      & 0.16    & \textbf{0.31}        \\
			\midrule
			\multirow{6}{*}{\textbf{\normalsize{Transfer}}} & Meta-Llama-3-8B-Instruct                     & 0.99                                             & 0.29                                  & 0.63      & 0.58    & 0.18 & 0.50 \\
			                                                & Gemma-2-9B-Instruct                          & 0.98                                             & 0.35                                  & 0.65      & 0.97    & 0.21 & 0.41 \\
			                                                & Meta-Llama-3-70B-Instruct                    & 0.97                                             & 0.73                                  & 0.93      & 1.00    & 0.38 & 0.75 \\
			                                                & GPT-3.5-turbo                                & 0.98                                             & 0.73                                  & 0.93      & 0.91    & 0.32 & 0.53 \\
			                                                & GPT-4o                                       & 0.98                                             & 0.61                                  & 0.88      & 0.95    & 0.25 & 0.53 \\
			\cmidrule(lr){2-8}
			                                                & \textit{Average}                                   & 0.98                                             & 0.54                                  & \textbf{0.80}      & 0.88    & 0.27 & \textbf{0.54} \\
			\bottomrule
		\end{tabular}}
	\label{tab:inference_on_ul}
\vspace{-0.1cm}
\end{table*}

\section{Unnatural Languages Can be Understood Across Tasks and LLMs}\label{sec:ul_is_comprehensible_by_llms}


In this section, we investigate whether unnatural languages—generated by
Algorithm \ref{alg:searching_ul} via reconstruction tasks across multiple
models—can be directly transferred across a broader range of tasks and
understood by diverse LLMs.

\subsection{Experiment Setup}

To evaluate LLMs' genuine understanding of unnatural languages, we design questions closely related to unnatural contexts. We employ context-based question-answering problems where the context is expressed in unnatural languages while maintaining questions in natural language. This approach helps isolate the models' understanding of unnatural languages without introducing additional comprehension challenges.

\noindent\textbf{Benchmarks.} \;
\ding{182} \emph{\SC}. We begin with a classic commonsense question-answering task, where questions are
asked in relation to given contextual knowledge. However, using existing
commonsense QA datasets presents a challenge, as LLMs may answer questions based
on their pre-trained knowledge rather than the provided context. To address
this, we leverage GPT-3.5~\citep{achiam2023gpt} to generate knowledge about
non-existing entities and their corresponding questions, named as \SC, thereby
ensuring the model must derive answers from the given context rather than rely
on pre-existing information. Prompts for generation and post-processing are
illustrated in Appendix \ref{sec:appen_prompt}. We then transformed natural
contexts into unnatural versions using the unnatural languages searching method
while preserving the original questions. \ding{183} \emph{\SG}. Furthermore, to ensure models do not
simply extract keywords from unnatural contexts in \SC, we test the unnatural
languages on GSM8K~\citep{cobbe2021training}, a more complex task requiring
reasoning capability. As our primary objective is to assess the ability to
comprehend unnatural languages, rather than to evaluate reasoning ability, we
select $100$ relatively simple questions from the whole test set, named \SG. We
show a concrete example for each dataset in Table \ref{table:dataset}.

\noindent\textbf{Backbone Models.} \;
We select a diverse range of LLMs, spanning from smaller open-source models to larger closed-source ones. Specifically, we use Mistral-7B-Instruct-v0.1~\citep{jiang2023mistral}, Vicuna-7B-v1.5 \citep{vicuna2023}, Meta-Llama-3-8B-Instruct~\citep{dubey2024llama}, Gemma2-9B-Instruct \citep{team2024gemma}, Meta-Llama-3-70B-Instruct, GPT-3.5-turbo~\citep{achiam2023gpt}, GPT-4o~\citep{hurst2024gpt}. Furthermore, to balance efficiency in the unnatural languages searching algorithm with generalizability, we employ Mistral-7B-Instruct-v0.1 and Vicuna-7B-v1.5, two renowned open-source models from distinct series, as the model set $M$ in Algorithm \ref{alg:searching_ul}.

\noindent\textbf{Baselines.} \;
We mainly employ two baselines. (i) natural language, which uses the original
unmodified text, and (ii) shuffled language with injected special tokens
(Shuf-Inj), which serves as the initialization step for our unnatural languages
search algorithm in Algorithm  \ref{alg:searching_ul}.

\begin{figure*}[t]
	\begin{center}
		\includegraphics[width=0.95\textwidth]{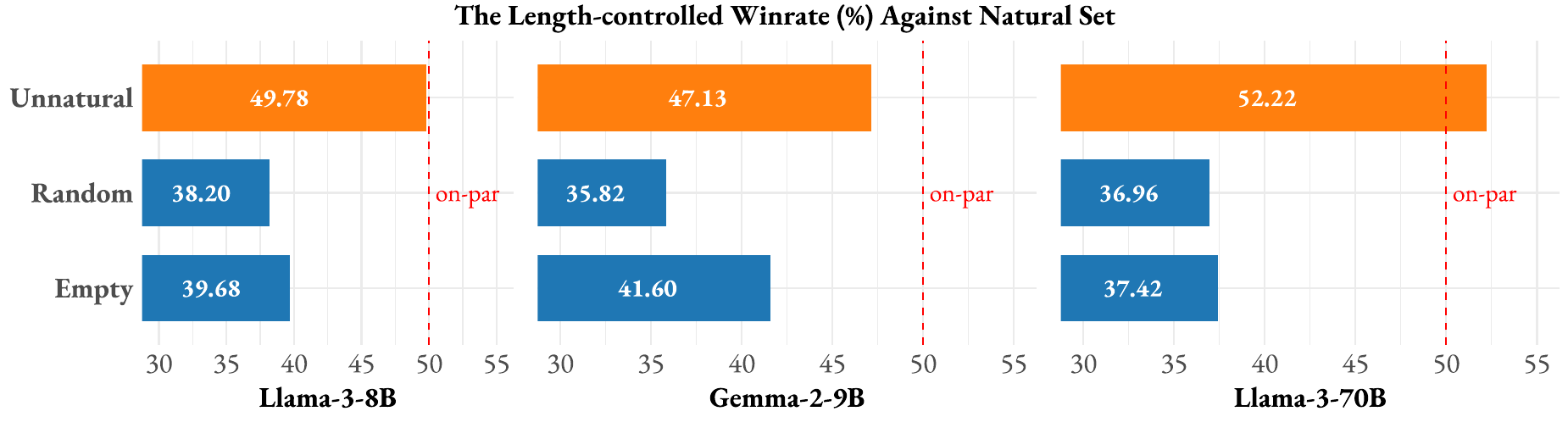}
	\end{center}
    \vspace{-5mm}
	\caption{The winrate (\%) against natural set on LC Alpaca 2.0 for various models. A winrate of 50\% indicates on-par
		performance.}\label{fig:alpacaeval_results}
\vspace{-0.2cm} 
\end{figure*}

\noindent\textbf{Experiment Details.} \;
For \SC, we evaluate performance using exact keyword matching, while for \SG, we use
accuracy as the evaluation metric.

\subsection{Main Results}

As shown in Table~\ref{tab:inference_on_ul}, for \SC, the test accuracy of all
models in unnatural languages is on-par with the one of natural language, by a
large margin with Shuf-InJ. In average, the test accuracy of transferred models
in unnatural languages is 80.4\%. For \SG, the test accuracy of most models on
natural questions is over 80\%, indicating the simplicity of questions, thus
mitigating the concerning of question complexity. Meanwhile, the performance of
close-source models could also answer half of the questions correctly,
outperforming the Shuf-InJ by an averaged margin of 26.7\%. Both results
indicate that such unnatural languages is highly transferrable across models with
different architectures and training corpus, including
GPT-4o~\citep{hurst2024gpt}, which is considered as the most well-aligned models.
As a result, it mitigates the conjecture that such unnatural languages is a
glitch of specific LLMs, but a general phenomenon and inherent property for
LLMs.

It is worthwhile to note that there is a significant performance gap between the
natural and unnatural set for \SG. This gap can be attributed to the increased
complexity of \SG contexts, which typically comprise multiple interconnected
sentences. Furthermore, \SG questions require sophisticated multi-step reasoning
processes, making them substantially more challenging than standard \SC tasks.
However, this is not the upper bound of the performance of LLMs
on unnatural languages, once the unnatural languages searching approach could be
further improved. 

In addition, we extend our investigation into the understanding of unnatural languages in a dialogue format, which serves as the foundation for LLM agents. Further details can be found in Appendix \ref{sec:appen_two}.

\subsection{Further Analysis}

To broaden the scope of the unnatural languages, we implemented it in the base
model rather than only the chat version, demonstrating that models can truly
understand these unnatural languages without relying on chat models' ability to
understand noisy instructions.

\noindent\textbf{Experiment Settings.} \;
For the base model, we employ in-context
learning (ICL) with examples in natural language to ensure consistent output
formatting while avoiding unnatural languages patterns in the learning process.

\noindent\textbf{Main Results.} \;
As shown in Table~\ref{tab:alig_pre_comparison}, under in-context learning setting with 8 examples,
the unnatural test accuracy of pre-trained base models before alignment achieves 38\% and 42\% in average, respectively. Particularly, considering the
success ratio (i.e. unnatural acc./natural acc.), the ratio achieves 53\% and
48\%, respectively. This indicates that
pre-trained model could inherently understand the unnatural languages without
alignment. 

\begin{table}[t]
	\caption{The performance comparison of different pre-trained base models on
		SimGSM8K. Ratio is calculated as the performance on unnatural languages divided by
		the performance on natural ones}\label{tab:alig_pre_comparison}
	\centering
	\setlength{\tabcolsep}{3pt}
	\scalebox{0.94}{
		\begin{tabular}{lc|cc|c}
			\toprule
			\textbf{Model}              & \textbf{Prompt} & \textbf{Natural} & \textbf{Unnatural} &
			\textbf{Ratio}                                                                               \\
			\midrule
			\multirow{2}{*}{Mistral-7B} & ICL(n=1)        & 0.70             & 0.23               & 0.33 \\
			                            & ICL(n=8)        & 0.71             & 0.38               & 0.53 \\
			\midrule
			\multirow{2}{*}{Llama-3-8B} & ICL(n=1)        & 0.74             & 0.33               & 0.45 \\
			                            & ICL(n=8)        & 0.87             & 0.42               & 0.48 \\
			\bottomrule
		\end{tabular}}
    \vspace{-0.3cm}
\end{table}

\section{LLMs Can Learn Instruction Following Capabilities From Unnatural Languages}\label{sec:ul_is_features}

\begin{table*}[t]
	\caption{The results of model variants instruction tuned on different types of \LIMA (including natural, unnatural, random, and empty instruction~\citep{hewitt2024instructionfollowinginstructiontuning})
		on MixEval~\citep{ni2024mixeval}. Each dataset represents the subsets selected by MixEval based on the real-world data distribution.
		M.C. and F.F. denote multiple-choice and free-form, respectively. Particularly, we
		remove the results of subset GPQA, MBPP, WinoGrande, and HumanEval since the subsets were too small (less than
		ten test cases) for consistent evaluation.}\label{tab:results_mixeval}
        \setlength{\tabcolsep}{2.4pt}
	\scalebox{0.9}{
		\begin{tabular}[c]{ll|>{\columncolor{lightgray}}ccc>{\columncolor{lightblue}}c|>{\columncolor{lightgray}}ccc>{\columncolor{lightblue}}c|>{\columncolor{lightgray}}ccc>{\columncolor{lightblue}}c}
			\toprule
			\multirow{2}{*}{\textbf{\normalsize{Type}}}          & \multirow{2}{*}{\textbf{\normalsize{Dataset}}}       &
			\multicolumn{4}{c}{\textbf{\normalsize{Llama-3-8B}}} \vline & \multicolumn{4}{c}{\textbf{\normalsize{Gemma-2-9B}}} \vline &
			\multicolumn{4}{c}{\textbf{\normalsize{Llama-3-70B}}}                                                                                                                                                                                                             \\
			\cmidrule(lr){3-6} \cmidrule(lr){7-10}      \cmidrule(lr){11-14}
			                                                     &                                                      &
			Natural                                              & Random
			                                                     & Empty                                                &
			Unnatural                                            & Natural
			                                                     & Random                                               &
			Empty                                                & Unnatural
			                                                     & Natural                                              &
			Random                                               & Empty                                                & Unnatural                                                                                                                                           \\
			\midrule
			\multirow{11}{*}{\textbf{\normalsize{M.C}}}          & ComsenseQA                                        & 0.530          & 0.569 & 0.604 & 0.579          & 0.599          & 0.550 & 0.495 & 0.668          & 0.748          & 0.668 & 0.559 & 0.693          \\
			                                                     & BoolQ                                                & 0.614          & 0.649 & 0.673 & 0.678          & 0.567          & 0.614 & 0.632 & 0.673          & 0.830          & 0.708 & 0.632 & 0.848          \\
			                                                     & OpenBookQA                                           & 0.581          & 0.628 & 0.721 & 0.721          & 0.605          & 0.744 & 0.744 & 0.721          & 0.814          & 0.744 & 0.721 & 0.837          \\
			                                                     & SIQA                                                 & 0.462          & 0.570 & 0.624 & 0.462          & 0.613          & 0.516 & 0.645 & 0.538          & 0.742          & 0.581 & 0.495 & 0.677          \\
			                                                     & HellaSwag                                            & 0.338          & 0.364 & 0.360 & 0.328          & 0.331          & 0.344 & 0.289 & 0.357          & 0.461          & 0.373 & 0.351 & 0.364          \\
			                                                     & MMLU-Pro                                             & 0.357          & 0.346 & 0.308 & 0.335          & 0.427          & 0.438 & 0.459 & 0.432          & 0.503          & 0.427 & 0.465 & 0.578          \\
			                                                     & AGIEval                                              & 0.331          & 0.359 & 0.340 & 0.352          & 0.370          & 0.314 & 0.407 & 0.349          & 0.566          & 0.423 & 0.426 & 0.543          \\
			                                                     & PIQA                                                 & 0.514          & 0.676 & 0.705 & 0.600          & 0.781          & 0.695 & 0.638 & 0.733          & 0.790          & 0.752 & 0.724 & 0.886          \\
			                                                     & MMLU                                                 & 0.658          & 0.634 & 0.661 & 0.633          & 0.724          & 0.680 & 0.700 & 0.718          & 0.811          & 0.736 & 0.731 & 0.805          \\
			                                                     & ARC                                                  & 0.802          & 0.802 & 0.780 & 0.791          & 0.923          & 0.901 & 0.879 & 0.923          & 0.956          & 0.835 & 0.868 & 0.934          \\\cmidrule(lr){2-14}
			
             & \textit{Average}                                                 & 0.545          & 0.563 & \textbf{0.579} & 0.552          & 0.607          & 0.585 & 0.582 & \textbf{0.623}          & 0.721          & 0.635 & 0.611 & \textbf{0.707}          \\\midrule
			\multirow{6}{*}{\textbf{\normalsize{F.F.}}}          & TriviaQA                                             & 0.591          & 0.453 & 0.452 & 0.558          & 0.609          & 0.481 & 0.563 & 0.585          & 0.829          & 0.638 & 0.685 & 0.825          \\
			                                                     & BBH                                                  & 0.537          & 0.633 & 0.526 & 0.606          & 0.621          & 0.438 & 0.700 & 0.687          & 0.817          & 0.670 & 0.484 & 0.693          \\
			                                                     & DROP                                                 & 0.584          & 0.385 & 0.484 & 0.545          & 0.638          & 0.481 & 0.638 & 0.651          & 0.755          & 0.585 & 0.631 & 0.767          \\
			                                                     & MATH                                                 & 0.381          & 0.290 & 0.510 & 0.394          & 0.490          & 0.668 & 0.568 & 0.410          & 0.668          & 0.642 & 0.606 & 0.610          \\
			                                                     & GSM8K                                                & 0.593          & 0.470 & 0.535 & 0.500          & 0.675          & 0.460 & 0.715 & 0.545          & 0.873          & 0.827 & 0.817 & 0.787          \\
			\cmidrule(lr){2-14}
			                                                    & \textit{Average}                                                 & 0.583          & 0.445 & 0.467 & \textbf{0.554}          & 0.616          & 0.481 & 0.592 & \textbf{0.603}          & 0.809          & 0.631 & 0.662 & \textbf{0.799}          \\\midrule
			                                                    
			                                                     \multicolumn{2}{c}{\textbf{Overall Average}} \vline                                            & 0.557 & 0.499 & 0.516 & \textbf{0.547} & 0.605 & 0.526 & 0.582 & \textbf{0.605} & 0.760 & 0.627 & 0.631 & \textbf{0.748} \\
			\bottomrule
		\end{tabular}
	}
\vspace{-0.3cm}
\end{table*}

In this section, we explore the properties of unnatural languages from the perspective of post-training.

\subsection{Experiment Setup}~\label{sec:it_dataset_and_experiemnt_settings}

\vspace{-0.5cm}

We explore whether the instruction fine-tuning
pre-trained LLMs on unnatural languages instructions could help models gain
general instruction following (chatting) ability.

\noindent\textbf{Training Dataset.} \;
We employ \LIMA~\citep{zhou2023lima}, a high-quality instruction tuning dataset
of $1000$ carefully created (instruction, answer) pairs. Furthermore, we
leverage our proposed unnatural languages searching approach to find an unnatural version for each
instruction in \LIMA, and keep the original answers in natural version.

\noindent\textbf{Benchmarks.} \;
We evaluate all variants on Length-controlled (LC)
AlpacaEval 2.0~\citep{alpaca_eval} and MixEval~\citep{ni2024mixeval}. LC
AlpacaEval 2.0 is a well-recognized benchmark for chat model
evaluation. MixEval is a ground-truth-based benchmark that collects
data from numerous QA datasets under real-world data distribution.

\noindent\textbf{Baselines.} \;
We employ three baselines: (i) Natural, which uses the original unmodified
instructions; (ii) Random, which replace the original instructions with an equal
number of random tokens; and (iii) Empty, in which instruction is empty~\citep{hewitt2024instructionfollowinginstructiontuning}.

\noindent\textbf{Experiment Details.} \; 
In practice, since the instruction in \LIMA
is extremely long, which exceeds the capacity of our searching approach, we
leverage GPT-4 to generate a compressed version of the instructions. Therefore,
for fairness, we compare the instruction following ability of models finetuned
on the unnatural \LIMA and the instruction-shortened \LIMA.  In addition, all
models are fine-tuned for $10$ epochs using identical hyperparameters.

\subsection{Main Results}

In Figure~\ref{fig:alpacaeval_results}, we show the winrate of different variants
against the corresponding models tuned on natural \LIMA using the official
pipeline and the annotation model is GPT-4o. The results clearly show that
responses of models instruction tuned on unnatural \LIMA is comparable to the
one tuned on natural \LIMA with a winrate of \(48.82\%\) in average, which
outperforms the baselines (models tuned on random/empty \LIMA) by a large
margin. 

Furthermore, Table~\ref{tab:results_mixeval} confirms our conclusion. As
shown in Table~\ref{tab:results_mixeval}, most models instruction tuned on
unnatural version \LIMA performs on-par with the one tuned on natural \LIMA.
Meanwhile, it outperforms the baselines, which are tuned on random and empty instruction version \LIMA, significantly, especially under base
model Llama-3-70B with a margin of over \(11\%\). Both results strongly
demonstrate our claim that unnatural languages contains generalizable patterns
that could be help LLMs gain instruction-following ability via instruction
finetuning.


\begin{figure*}[t]
    \centering
    \begin{minipage}{0.4\textwidth}
        \centering
        \captionof{table}{The accuracy of the unnatural GSM8K test set for models instruction-tuned on various types of GSM8K training sets. Columns denote the training set.}
        \label{tab:unnatural_gsm8k}
        \setlength{\tabcolsep}{3pt}
        \scalebox{0.76}{
        \begin{tabular}{l|ccc|c}
            \toprule
            \textbf{Unnatural Test Acc.} & \textbf{Natural} & \textbf{Random} & \textbf{Empty} & \textbf{Unnatural} \\
            \midrule
            Mistral-7B                  & 0.187            & 0.116           & 0.122          & \textbf{0.310} \\
            Llama-3-8B                  & 0.197            & 0.087           & 0.109          & \textbf{0.312} \\
            Mistral-7B-Instruct         & 0.225            & 0.138           & 0.171          & \textbf{0.300} \\
            Llama-3-8B-Instruct         & 0.214            & 0.136           & 0.179          & \textbf{0.349} \\
            \midrule
            \textit{Average}                     & 0.206            & 0.119           & 0.145          & \textbf{0.318} \\
            \bottomrule
        \end{tabular}}
    \end{minipage}
    \hfill
    \begin{minipage}{0.56\textwidth}
        \centering
 \begin{subfigure}[b]{0.49\textwidth}
            \includegraphics[width=\textwidth]{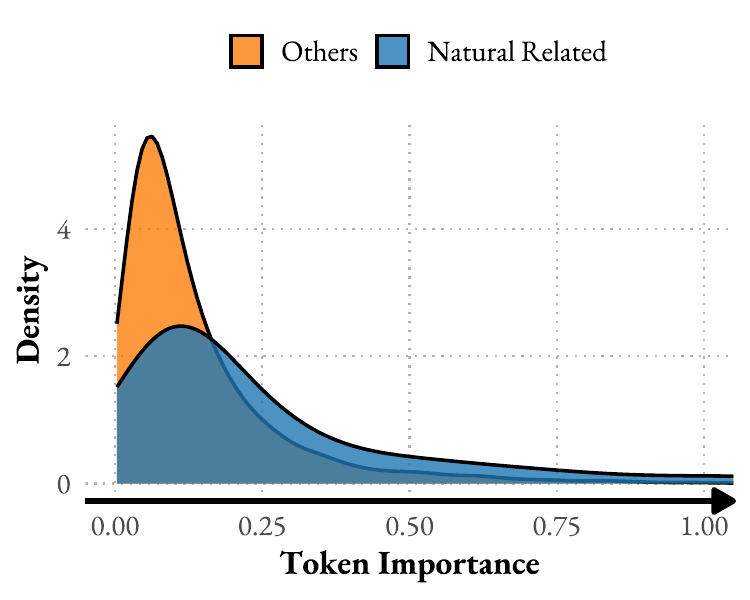}
            \centering
            \caption{Token importance distribution}
        \end{subfigure}
        \hfill
        \begin{subfigure}[b]{0.49\textwidth}
            \includegraphics[width=\textwidth]{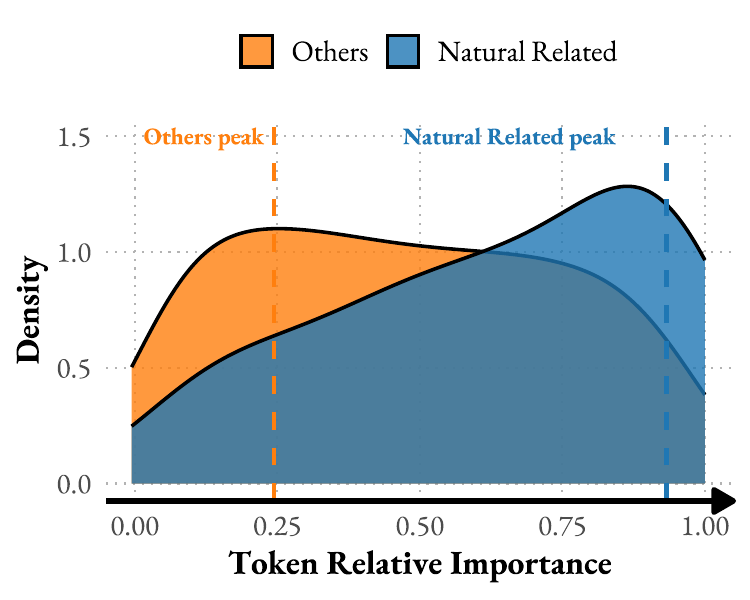}
            \caption{Relative importance distribution}
        \end{subfigure}
        \centering
        \caption{Token importance of \SG under Llama3-8B-Instruct.}
        \label{fig:tok_imp_dist}
    \end{minipage}
\end{figure*}

\subsection{Further Analysis}\label{sec:unnatural_patterns}


So far we have shown that unnatural languages contains natural patterns that
could be generalized to various tasks by instruction tuning. Besides, we are
curious about whether the unnatural languages consists unnatural
patterns and whether fine-tuning on unnatural languages could boost the unnatural
language understanding capability of LLMs. To this end, we focus the math
reasoning task, which is highly demanded for question understanding. We created
an unnatural languages version of GSM8K for its training subset and test subset.
Due to the high cost of GCG and computation limitation, we searched \(1333\)
training instances and \(654\) test instances. For training, we leverage
corresponding answer augmentation version (i.e., for each question, there are
multiple version of correct chain-of-thought answers.) from \citet{yu2023metamath}, which
finally results in \(14886\) training instances. Built upon the training set, we
create four versions, the same as Section~\ref{sec:it_dataset_and_experiemnt_settings}.
We train the pre-trained version and instruction tuned version of
Mistral-7B-v0.1 and Llama-3-8B on the four types of training
set and test their performance on the unnatural test set. The results are shown
in Table~\ref{tab:unnatural_gsm8k}.

It shows that models fine-tuned on the unnatural training set significantly outperform models trained on other types of training sets when evaluated on the unnatural test set. Specifically, the average accuracy for
models tuned on unnatural training set is \(31.8\%\), outperforming the one tuned on natural training set by \(11.2\%\),
the one tuned on random training set by \(19.9\%\), and the one tuned on empty training set by \(17.3\%\). This
indicates that unnatural languages contain generalizable unnatural patterns that could enhance
LLMs' unnatural languages understanding capability.

\section{How do LLMs Understand Unnatural Languages?}

Unnatural languages have been empirically shown to contain latent features that are comprehensible across different LLMs, while also enhancing their ability to follow instructions. In this section, we investigate how LLMs process and understand these unnatural languages.

\begin{figure*}[t]
	\centering
	\begin{subfigure}[b]{0.33\textwidth}
		\centering
		\includegraphics[width=\textwidth]{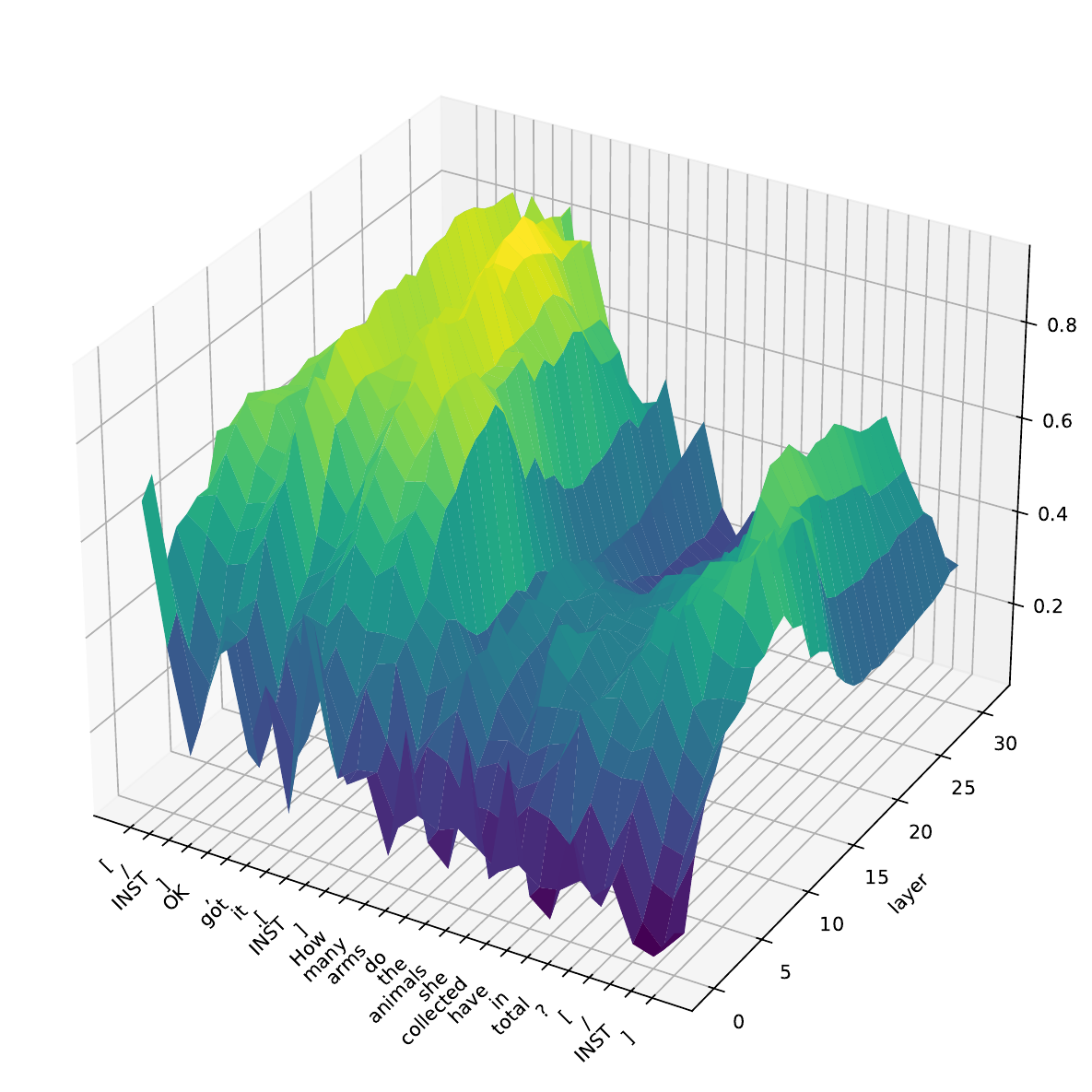}
	\end{subfigure}
	\begin{subfigure}[b]{0.33\textwidth}
		\centering
		\includegraphics[width=\textwidth]{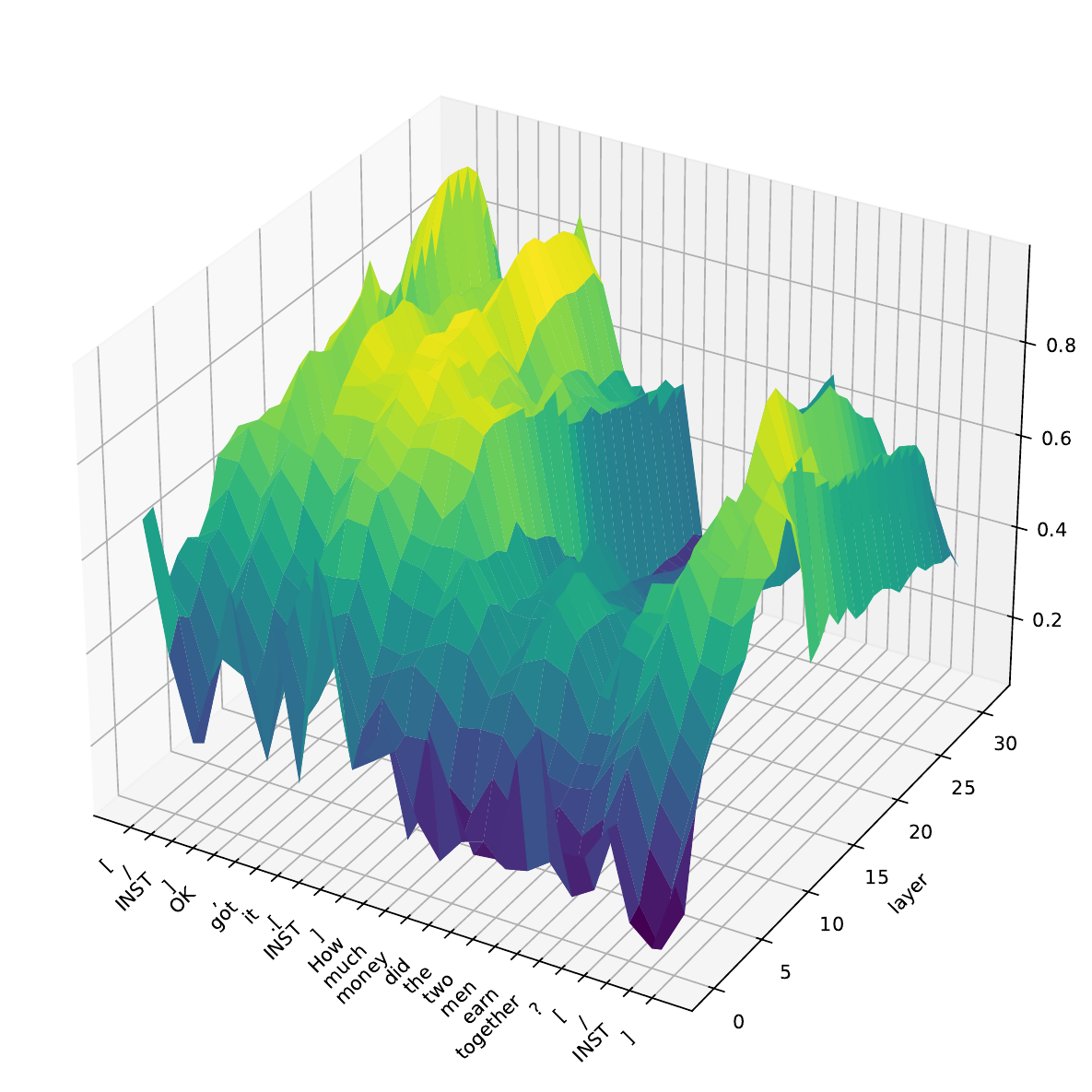}
	\end{subfigure}
	\begin{subfigure}[b]{0.33\textwidth}
		\centering
		\includegraphics[width=\textwidth]{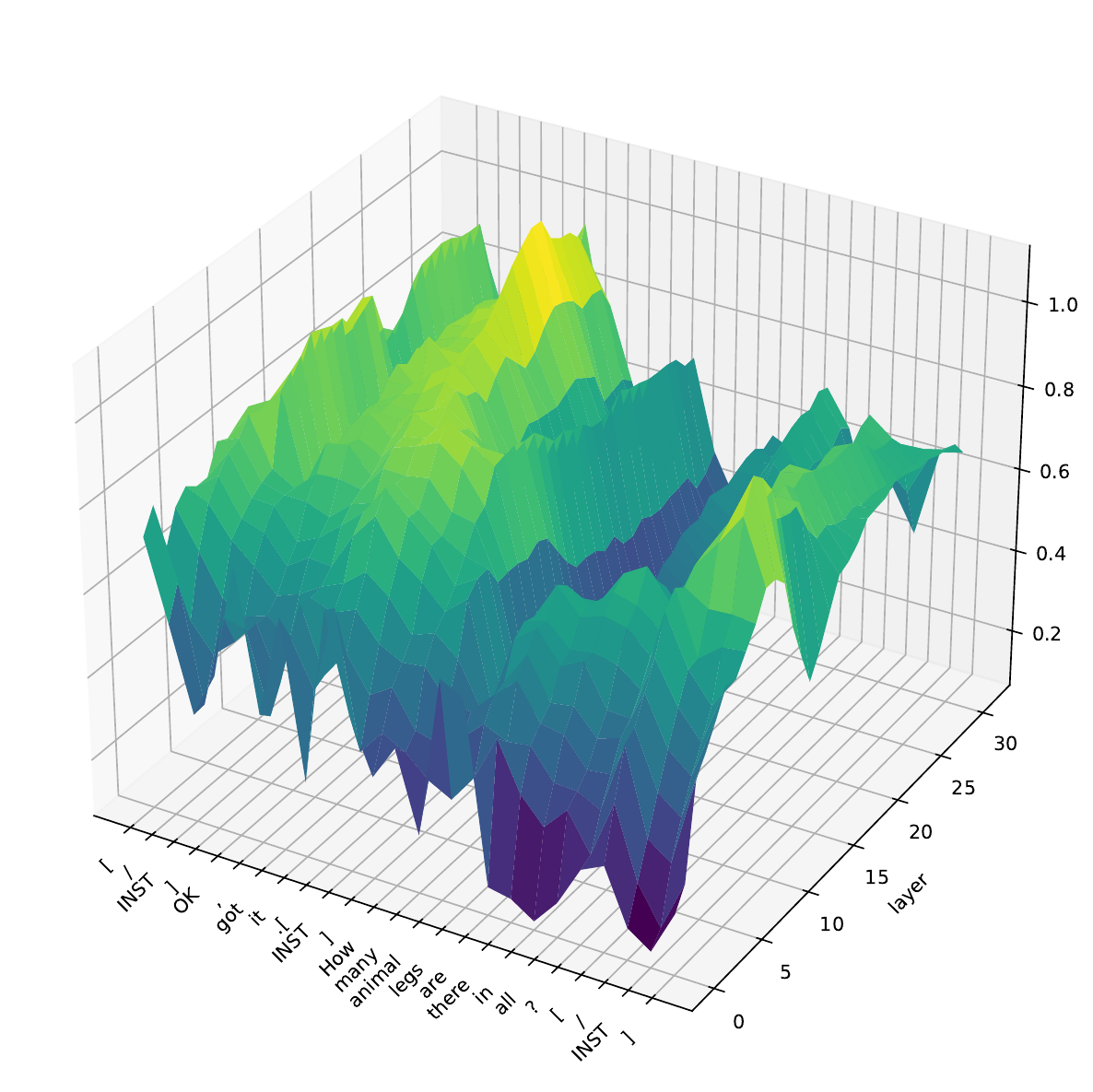}
	\end{subfigure}
	\caption{Three 3D surface examples showing the inverse similarity of natural and unnatural context
		embeddings. The inverse similarity decreases significantly when question-related tokens are added, indicating that LLMs correctly infers the organization of keywords.}\label{fig:sim_cee}
\vspace{-0.2cm}
\end{figure*}

\subsection{LLMs Extract Keywords from Unnatural Languages}

To investigate what LLMs truly capture when processing unnatural languages, we evaluate each token's importance by measuring its impact on the output when removed from the sequence. Formally, for an unnatural string $S'$, tokenized by model $M$ as $\mathbf{x}_{1:n} = [x_1, x_2, \cdots, x_n]$, the importance of token $x_i$ is defined as the effect of its removal on the change in the embedding of $M$'s final layer, i.e.,
\begin{equation}
    I(x_i) = \Big\|\mathcal{LLM}_M(\mathbf{x}_{1:n}) - \mathcal{LLM}_M(\mathbf{x}_{1:n} \textbackslash x_i)\Big\|_2.
\end{equation}
Specifically, to ensure consistent evaluation across unnatural strings of varying lengths and given that the embedding of the final position is used to predict the next token, we measure the embedding of the final position rather than the entire sequence. Furthermore, we normalize the ``token importance'' relative to the most important token in each sequence, referring to this as ``relative token importance''. The relative token importance represents a softened position of the token after sorting in ascending order, scaled within the range \([0, 1]\). Tokens with greater importance within a data point have a relative position closer to $1$, whereas less important tokens have a relative position closer to $0$.

In Figure \ref{fig:tok_imp_dist}, we present the distribution of token importance and relative token importance within the unnatural version of the SimGSM8K dataset as processed by the model Llama-3-8B-Instruct. Specifically, as shown in the Figure~\ref{fig:tok_imp_dist} (a), the density of natural-related tokens (tokens appears in the natural version) is more higher than others
when token importance is higher than 0.2. Besides, compared with natural related tokens,
the majority of other tokens lies in the lower importance range. Furthermore, Figure~\ref{fig:tok_imp_dist} (b) clearly shows that most of the naturally related tokens are higher relative important while the
other tokens are lower relative important. The above results demonstrate that LLMs
are capable of pay more attention on the natural related tokens while filtering out the other noise. Consequently, LLMs effectively extract keywords from unnatural languages inputs.

\subsection{LLMs Infer Correct Organization of Keywords in Unnatural Languages}\label{sec:llm_understand_ul_wrt_context}

\begin{table*}[t]
	\caption{Concrete examples of token reordering: The ``natural context'' represents the original version, while the ``unnatural version'' simplifies the unnatural languages by removing noise and retaining only keywords. We decode the internal embeddings of LLMs into tokens for unnatural language inputs using the same decoder as the final output layer, referred to as ``decode internal embeddings''.}
	\centering
	\scalebox{0.9}{
		\begin{tabular}{p{6cm}|p{5cm}|p{6cm}}
			\toprule
			 \textbf{Natural Context} & \textbf{Unnatural Version (De-noised)}  & \textbf{Decode Internal Embeddings} \\
             \midrule
             Brandon sold 86 geckos last year. He sold twice that many the year before. & twice geckos year before last sold Brandon He & twice geckos year before last before sold \colorbox{lightblue}{Brandon He sold} he \colorbox{lightblue}{86} sold twice sold twice sold \colorbox{lightblue}{last} twice \colorbox{lightblue}{year} sold \\
             \midrule
             Ruiz receives a monthly salary of \$500. & a monthly \$500 salary Ruiz receives &  a \colorbox{lightblue}{monthly \$500} a salary a \colorbox{lightblue}{Ruiz} a receives a \colorbox{lightblue}{receives a salary}\\
			\bottomrule
		\end{tabular}}
	\label{table:reorder_example}
\vspace{-0.2cm}
\end{table*}

Although LLMs are capable of extracting keywords from unnatural languages, the extracted words are often shuffled and arranged in the wrong order. Therefore, we hypothesize that LLMs can reorganize these keywords and infer their correct arrangement. Specifically, when the unnatural context is provided to LLMs alongside appended natural questions, we propose that LLMs progressively infer the correct organization of keywords—i.e., their corresponding natural language versions—as they process an increasing number of natural question tokens.

To verify our hypothesis, we calculate the inverse similarity of embeddings between the context inputs of unnatural languages and their corresponding natural versions across layers, as well as for increasingly natural question tokens inputted into LLMs. As shown in Figure~\ref{fig:sim_cee}, for the marginal version of a layer, the inverse similarity of
unnatural and natural context embeddings gradually decreases as following tokens are inputted. Particularly,
the inverse similarity drastically decreases when the question related tokens are inputted.
This indicates that LLMs does not always comprehend unnatural languages independently as the corresponding natural
language. In contrast, the comprehension process is highly related to the context (i.e. the questions.).
As a result, the unnatural languages works in certain contexts.
Supposing random context is provided, the model behavior on unnatural languages could be different from natural language.

\subsection{Qualitative Analysis}

We further investigate whether LLMs are truly capable of reordering keywords by analyzing the embeddings of intermediate layers. Specifically, we decode the internal embeddings of LLMs when processing unnatural language inputs into tokens using the same decoder as the final layer. As shown in Table \ref{table:reorder_example}, we observe that although the keywords in the unnatural version are disordered, LLMs are able to reorder certain patterns in the keywords to match the original natural context. Moreover, we utilize dependency parsing to demonstrate that LLMs can understand the dependency structure of unnatural languages. Details are illustrated in Appendix \ref{sec:appen_example}.

\section{Related Works}

\noindent
\textbf{Unnatural Languages.} \;
Prior studies have observed isolated instances of unexpected model behavior, while they did not explicitly identify or systematically analyze unnatural language. For example, \citet{zou2023universal} prompted models generating harmful outputs, \citet{pfau2024let} enhanced chain-of-thought reasoning, and \citet{sinha2020unnatural} showed NLI models worked with permuted inputs. \citet{kervadec2023unnatural} demonstrated that LLMs interpret unnatural languages differently, while \citet{kallini2024mission} categorized them by perplexity, showing LLMs struggled to learn them. 

\noindent
\textbf{Discrete Optimization.} \;
Prompt optimization tools have been used to explore unnatural languages in token space. Early works~\citep{zou2023universal,liu2023autodan,zhao2024accelerating,chao2023jailbreaking,andriushchenko2022adversarial} focused on adversarial prompts to jailbreak LLMs, while others~\citep{shin2020autoprompt,jones2023automatically} optimized prompts for specific outputs. These studies did not address whether such prompts reflect natural language features. This work investigates this question.

\noindent
\textbf{Transferability of Adversarial Examples.} \;
Unnatural languages often transfer across LLMs, similar to adversarial examples in computer vision~\cite{szegedy2014intriguing,papernot2016transferability,nguyen2015deep}. For example, \citet{wallace2019universal} showed that prompts from GPT-2 transferred to larger models, and \citet{jones2023automatically} found that toxic prompts from GPT-2 affected davinci-002. Building on these findings, this work examines how LLMs rely on fragile, unnatural features in token space~\citep{ilyas2019adversarial}.

\section{Conclusion}

Our study reveals that LLMs possess the ability to comprehend unnatural languages,
an incomprehensible data pattern that could convey information across models.
Through systematic analysis and experiments, we demonstrate that unnatural languages
contains generalizable patterns across a wide variety of LLMs, despite these models being
predominantly aligned with human data. We show that models fine-tuned on unnatural instructions achieves on-par performance of instruction-following ability with models fine-tuned on
natural versions. Furthermore, through comprehensive analysis, we demonstrate that LLMs process unnatural languages by filtering noise and inferring contextual meaning from filtered words.

\section*{Impact Statement}
Our findings suggest that LLMs' interaction with incomprehensible patterns, while potentially
useful, adds complexity to our understanding of their behavior and raises important questions about the nature of
language processing in artificial systems. This could have broader implications for the development and deployment of AI technologies. Understanding these dynamics is crucial for ensuring the responsible use of LLMs in real-world applications, where their ability to process unconventional data patterns might lead to both innovative solutions and unforeseen challenges.


\section*{Acknowledgment}
We thank Wei Tsang Ooi, Ying Sheng, Lianmin Zheng, Hwee Tou Ng, Tat-Seng Chua, and Dan Hendrycks for their insightful discussions and feedback. We are also grateful to the Center for AI Safety (CAIS) for providing computational resources that supported this research.

\bibliography{reference}

\begin{thebibliography}{41}
\providecommand{\natexlab}[1]{#1}
\providecommand{\url}[1]{\texttt{#1}}
\expandafter\ifx\csname urlstyle\endcsname\relax
  \providecommand{\doi}[1]{doi: #1}\else
  \providecommand{\doi}{doi: \begingroup \urlstyle{rm}\Url}\fi

\bibitem[Achiam et~al.(2023)Achiam, Adler, Agarwal, Ahmad, Akkaya, Aleman,
  Almeida, Altenschmidt, Altman, Anadkat, et~al.]{achiam2023gpt}
Achiam, J., Adler, S., Agarwal, S., Ahmad, L., Akkaya, I., Aleman, F.~L.,
  Almeida, D., Altenschmidt, J., Altman, S., Anadkat, S., et~al.
\newblock Gpt-4 technical report.
\newblock \emph{arXiv preprint arXiv:2303.08774}, 2023.

\bibitem[Andriushchenko(2022)]{andriushchenko2022adversarial}
Andriushchenko, M.
\newblock Adversarial attacks on gpt-4 via simple random search, 2022.

\bibitem[anthropic(2024)]{anthropic2024claude}
anthropic.
\newblock Meet claude.
\newblock \url{https://www.anthropic.com/claude}, 2024.
\newblock Accessed: 2024-05-01.

\bibitem[Bisk et~al.(2020)Bisk, Zellers, Gao, Choi, et~al.]{bisk2020piqa}
Bisk, Y., Zellers, R., Gao, J., Choi, Y., et~al.
\newblock Piqa: Reasoning about physical commonsense in natural language.
\newblock In \emph{Proceedings of the AAAI conference on artificial
  intelligence}, volume~34, pp.\  7432--7439, 2020.

\bibitem[Chao et~al.(2023)Chao, Robey, Dobriban, Hassani, Pappas, and
  Wong]{chao2023jailbreaking}
Chao, P., Robey, A., Dobriban, E., Hassani, H., Pappas, G.~J., and Wong, E.
\newblock Jailbreaking black box large language models in twenty queries.
\newblock \emph{arXiv preprint arXiv:2310.08419}, 2023.

\bibitem[Chiang et~al.(2023)Chiang, Li, Lin, Sheng, Wu, Zhang, Zheng, Zhuang,
  Zhuang, Gonzalez, Stoica, and Xing]{vicuna2023}
Chiang, W.-L., Li, Z., Lin, Z., Sheng, Y., Wu, Z., Zhang, H., Zheng, L.,
  Zhuang, S., Zhuang, Y., Gonzalez, J.~E., Stoica, I., and Xing, E.~P.
\newblock Vicuna: An open-source chatbot impressing gpt-4 with 90\%* chatgpt
  quality, March 2023.
\newblock URL \url{https://lmsys.org/blog/2023-03-30-vicuna/}.

\bibitem[Cobbe et~al.(2021)Cobbe, Kosaraju, Bavarian, Chen, Jun, Kaiser,
  Plappert, Tworek, Hilton, Nakano, et~al.]{cobbe2021training}
Cobbe, K., Kosaraju, V., Bavarian, M., Chen, M., Jun, H., Kaiser, L., Plappert,
  M., Tworek, J., Hilton, J., Nakano, R., et~al.
\newblock Training verifiers to solve math word problems, 2021.
\newblock \emph{URL https://arxiv. org/abs/2110.14168}, 2021.

\bibitem[Devlin(2018)]{devlin2018bert}
Devlin, J.
\newblock Bert: Pre-training of deep bidirectional transformers for language
  understanding.
\newblock \emph{arXiv preprint arXiv:1810.04805}, 2018.

\bibitem[Dubey et~al.(2024)Dubey, Jauhri, Pandey, Kadian, Al-Dahle, Letman,
  Mathur, Schelten, Yang, Fan, et~al.]{dubey2024llama}
Dubey, A., Jauhri, A., Pandey, A., Kadian, A., Al-Dahle, A., Letman, A.,
  Mathur, A., Schelten, A., Yang, A., Fan, A., et~al.
\newblock The llama 3 herd of models.
\newblock \emph{arXiv preprint arXiv:2407.21783}, 2024.

\bibitem[Gao et~al.(2024)Gao, Song, Yang, Cai, Miao, Dong, Li, Ma, Chen, Xu,
  et~al.]{gao2024omni}
Gao, B., Song, F., Yang, Z., Cai, Z., Miao, Y., Dong, Q., Li, L., Ma, C., Chen,
  L., Xu, R., et~al.
\newblock Omni-math: A universal olympiad level mathematic benchmark for large
  language models.
\newblock \emph{arXiv preprint arXiv:2410.07985}, 2024.

\bibitem[Hendrycks et~al.(2021)Hendrycks, Burns, Kadavath, Arora, Basart, Tang,
  Song, and Steinhardt]{hendrycks2021measuring}
Hendrycks, D., Burns, C., Kadavath, S., Arora, A., Basart, S., Tang, E., Song,
  D., and Steinhardt, J.
\newblock Measuring mathematical problem solving with the math dataset.
\newblock \emph{arXiv preprint arXiv:2103.03874}, 2021.

\bibitem[Hewitt \& Manning(2019)Hewitt and Manning]{hewitt2019structural}
Hewitt, J. and Manning, C.~D.
\newblock A structural probe for finding syntax in word representations.
\newblock In \emph{North American Chapter of the Association for Computational
  Linguistics: Human Language Technologies}. Association for Computational
  Linguistics, 2019.

\bibitem[Hewitt et~al.(2024)Hewitt, Liu, Liang, and
  Manning]{hewitt2024instructionfollowinginstructiontuning}
Hewitt, J., Liu, N.~F., Liang, P., and Manning, C.~D.
\newblock Instruction following without instruction tuning, 2024.
\newblock URL \url{https://arxiv.org/abs/2409.14254}.

\bibitem[Hurst et~al.(2024)Hurst, Lerer, Goucher, Perelman, Ramesh, Clark,
  Ostrow, Welihinda, Hayes, Radford, et~al.]{hurst2024gpt}
Hurst, A., Lerer, A., Goucher, A.~P., Perelman, A., Ramesh, A., Clark, A.,
  Ostrow, A., Welihinda, A., Hayes, A., Radford, A., et~al.
\newblock Gpt-4o system card.
\newblock \emph{arXiv preprint arXiv:2410.21276}, 2024.

\bibitem[Ilyas et~al.(2019)Ilyas, Santurkar, Tsipras, Engstrom, Tran, and
  Madry]{ilyas2019adversarial}
Ilyas, A., Santurkar, S., Tsipras, D., Engstrom, L., Tran, B., and Madry, A.
\newblock Adversarial examples are not bugs, they are features.
\newblock \emph{Advances in neural information processing systems}, 32, 2019.

\bibitem[Jiang et~al.(2023)Jiang, Sablayrolles, Mensch, Bamford, Chaplot,
  Casas, Bressand, Lengyel, Lample, Saulnier, et~al.]{jiang2023mistral}
Jiang, A.~Q., Sablayrolles, A., Mensch, A., Bamford, C., Chaplot, D.~S., Casas,
  D. d.~l., Bressand, F., Lengyel, G., Lample, G., Saulnier, L., et~al.
\newblock Mistral 7b.
\newblock \emph{arXiv preprint arXiv:2310.06825}, 2023.

\bibitem[Jones et~al.(2023)Jones, Dragan, Raghunathan, and
  Steinhardt]{jones2023automatically}
Jones, E., Dragan, A., Raghunathan, A., and Steinhardt, J.
\newblock Automatically auditing large language models via discrete
  optimization.
\newblock \emph{arXiv preprint arXiv:2303.04381}, 2023.

\bibitem[Kallini et~al.(2024)Kallini, Papadimitriou, Futrell, Mahowald, and
  Potts]{kallini2024mission}
Kallini, J., Papadimitriou, I., Futrell, R., Mahowald, K., and Potts, C.
\newblock Mission: Impossible language models.
\newblock \emph{arXiv preprint arXiv:2401.06416}, 2024.

\bibitem[Kervadec et~al.(2023)Kervadec, Franzon, and
  Baroni]{kervadec2023unnatural}
Kervadec, C., Franzon, F., and Baroni, M.
\newblock Unnatural language processing: How do language models handle
  machine-generated prompts?
\newblock \emph{arXiv preprint arXiv:2310.15829}, 2023.

\bibitem[Li et~al.(2023)Li, Zhang, Dubois, Taori, Gulrajani, Guestrin, Liang,
  and Hashimoto]{alpaca_eval}
Li, X., Zhang, T., Dubois, Y., Taori, R., Gulrajani, I., Guestrin, C., Liang,
  P., and Hashimoto, T.~B.
\newblock Alpacaeval: An automatic evaluator of instruction-following models.
\newblock \url{https://github.com/tatsu-lab/alpaca_eval}, 5 2023.

\bibitem[Liu et~al.(2023)Liu, Xu, Chen, and Xiao]{liu2023autodan}
Liu, X., Xu, N., Chen, M., and Xiao, C.
\newblock Autodan: Generating stealthy jailbreak prompts on aligned large
  language models.
\newblock \emph{arXiv preprint arXiv:2310.04451}, 2023.

\bibitem[Mundhenk et~al.(2020)Mundhenk, Chen, and
  Friedland]{mundhenk2020saliency}
Mundhenk, T.~N., Chen, B.~Y., and Friedland, G.
\newblock Efficient saliency maps for explainable ai, 2020.
\newblock URL \url{https://arxiv.org/abs/1911.11293}.

\bibitem[Nguyen et~al.(2015)Nguyen, Yosinski, and Clune]{nguyen2015deep}
Nguyen, A., Yosinski, J., and Clune, J.
\newblock Deep neural networks are easily fooled: High confidence predictions
  for unrecognizable images.
\newblock In \emph{Proceedings of the IEEE conference on computer vision and
  pattern recognition}, pp.\  427--436, 2015.

\bibitem[Ni et~al.(2024)Ni, Xue, Yue, Deng, Shah, Jain, Neubig, and
  You]{ni2024mixeval}
Ni, J., Xue, F., Yue, X., Deng, Y., Shah, M., Jain, K., Neubig, G., and You, Y.
\newblock Mixeval: Deriving wisdom of the crowd from llm benchmark mixtures.
\newblock \emph{arXiv preprint arXiv:2406.06565}, 2024.

\bibitem[OpenAI(2023)]{john2023chatgpt}
OpenAI.
\newblock Introducing chatgpt.
\newblock \url{https://openai.com/blog/chatgpt}, 2023.
\newblock Accessed: 2024-05-01.

\bibitem[Ouyang et~al.(2022)Ouyang, Wu, Jiang, Almeida, Wainwright, Mishkin,
  Zhang, Agarwal, Slama, Ray, et~al.]{ouyang2022training}
Ouyang, L., Wu, J., Jiang, X., Almeida, D., Wainwright, C., Mishkin, P., Zhang,
  C., Agarwal, S., Slama, K., Ray, A., et~al.
\newblock Training language models to follow instructions with human feedback.
\newblock \emph{Advances in Neural Information Processing Systems},
  35:\penalty0 27730--27744, 2022.

\bibitem[Papernot et~al.(2016)Papernot, McDaniel, and
  Goodfellow]{papernot2016transferability}
Papernot, N., McDaniel, P., and Goodfellow, I.
\newblock Transferability in machine learning: from phenomena to black-box
  attacks using adversarial samples.
\newblock \emph{arXiv preprint arXiv:1605.07277}, 2016.

\bibitem[Pfau et~al.(2024)Pfau, Merrill, and Bowman]{pfau2024let}
Pfau, J., Merrill, W., and Bowman, S.~R.
\newblock Let's think dot by dot: Hidden computation in transformer language
  models.
\newblock \emph{arXiv preprint arXiv:2404.15758}, 2024.

\bibitem[Reimers \& Gurevych(2019)Reimers and Gurevych]{reimers2019sentence}
Reimers, N. and Gurevych, I.
\newblock Sentence-bert: Sentence embeddings using siamese bert-networks.
\newblock \emph{arXiv preprint arXiv:1908.10084}, 2019.

\bibitem[Shin et~al.(2020)Shin, Razeghi, Logan~IV, Wallace, and
  Singh]{shin2020autoprompt}
Shin, T., Razeghi, Y., Logan~IV, R.~L., Wallace, E., and Singh, S.
\newblock Autoprompt: Eliciting knowledge from language models with
  automatically generated prompts.
\newblock \emph{arXiv preprint arXiv:2010.15980}, 2020.

\bibitem[Silveira et~al.(2014)Silveira, Dozat, de~Marneffe, Bowman, Connor,
  Bauer, and Manning]{silveira14gold}
Silveira, N., Dozat, T., de~Marneffe, M.-C., Bowman, S., Connor, M., Bauer, J.,
  and Manning, C.~D.
\newblock A gold standard dependency corpus for {E}nglish.
\newblock In \emph{Proceedings of the Ninth International Conference on
  Language Resources and Evaluation (LREC-2014)}, 2014.

\bibitem[Sinha et~al.(2020)Sinha, Parthasarathi, Pineau, and
  Williams]{sinha2020unnatural}
Sinha, K., Parthasarathi, P., Pineau, J., and Williams, A.
\newblock Unnatural language inference.
\newblock \emph{arXiv preprint arXiv:2101.00010}, 2020.

\bibitem[Szegedy et~al.(2014)Szegedy, Zaremba, Sutskever, Bruna, Erhan,
  Goodfellow, and Fergus]{szegedy2014intriguing}
Szegedy, C., Zaremba, W., Sutskever, I., Bruna, J., Erhan, D., Goodfellow,
  I.~J., and Fergus, R.
\newblock Intriguing properties of neural networks.
\newblock In \emph{International Conference on Learning Representations}, 2014.

\bibitem[Team et~al.(2024)Team, Riviere, Pathak, Sessa, Hardin, Bhupatiraju,
  Hussenot, Mesnard, Shahriari, Ram{\'e}, et~al.]{team2024gemma}
Team, G., Riviere, M., Pathak, S., Sessa, P.~G., Hardin, C., Bhupatiraju, S.,
  Hussenot, L., Mesnard, T., Shahriari, B., Ram{\'e}, A., et~al.
\newblock Gemma 2: Improving open language models at a practical size.
\newblock \emph{arXiv preprint arXiv:2408.00118}, 2024.

\bibitem[Touvron et~al.(2023)Touvron, Martin, Stone, Albert, Almahairi, Babaei,
  Bashlykov, Batra, Bhargava, Bhosale, et~al.]{touvron2023llama}
Touvron, H., Martin, L., Stone, K., Albert, P., Almahairi, A., Babaei, Y.,
  Bashlykov, N., Batra, S., Bhargava, P., Bhosale, S., et~al.
\newblock Llama 2: Open foundation and fine-tuned chat models.
\newblock \emph{arXiv preprint arXiv:2307.09288}, 2023.

\bibitem[Wallace et~al.(2019)Wallace, Feng, Kandpal, Gardner, and
  Singh]{wallace2019universal}
Wallace, E., Feng, S., Kandpal, N., Gardner, M., and Singh, S.
\newblock Universal adversarial triggers for attacking and analyzing nlp.
\newblock \emph{arXiv preprint arXiv:1908.07125}, 2019.

\bibitem[Wei et~al.(2021)Wei, Bosma, Zhao, Guu, Yu, Lester, Du, Dai, and
  Le]{wei2021finetuned}
Wei, J., Bosma, M., Zhao, V.~Y., Guu, K., Yu, A.~W., Lester, B., Du, N., Dai,
  A.~M., and Le, Q.~V.
\newblock Finetuned language models are zero-shot learners.
\newblock \emph{arXiv preprint arXiv:2109.01652}, 2021.

\bibitem[Yu et~al.(2023)Yu, Jiang, Shi, Yu, Liu, Zhang, Kwok, Li, Weller, and
  Liu]{yu2023metamath}
Yu, L., Jiang, W., Shi, H., Yu, J., Liu, Z., Zhang, Y., Kwok, J.~T., Li, Z.,
  Weller, A., and Liu, W.
\newblock Metamath: Bootstrap your own mathematical questions for large
  language models.
\newblock \emph{arXiv preprint arXiv:2309.12284}, 2023.

\bibitem[Zhao et~al.(2024)Zhao, Zheng, Cai, Do, Kawaguchi, Goyal, and
  Shieh]{zhao2024accelerating}
Zhao, Y., Zheng, W., Cai, T., Do, X.~L., Kawaguchi, K., Goyal, A., and Shieh,
  M.
\newblock Accelerating greedy coordinate gradient via probe sampling.
\newblock \emph{arXiv preprint arXiv:2403.01251}, 2024.

\bibitem[Zhou et~al.(2023)Zhou, Liu, Xu, Iyer, Sun, Mao, Ma, Efrat, Yu, Yu,
  et~al.]{zhou2023lima}
Zhou, C., Liu, P., Xu, P., Iyer, S., Sun, J., Mao, Y., Ma, X., Efrat, A., Yu,
  P., Yu, L., et~al.
\newblock Lima: Less is more for alignment.
\newblock \emph{arXiv preprint arXiv:2305.11206}, 2023.

\bibitem[Zou et~al.(2023)Zou, Wang, Kolter, and Fredrikson]{zou2023universal}
Zou, A., Wang, Z., Kolter, J.~Z., and Fredrikson, M.
\newblock Universal and transferable adversarial attacks on aligned language
  models.
\newblock \emph{arXiv preprint arXiv:2307.15043}, 2023.

\end{thebibliography}
\bibliographystyle{icml2025}

\onecolumn
\appendix
\renewcommand{\thepage}{A\arabic{page}}
\renewcommand{\thesection}{A\arabic{section}}
\renewcommand{\thetable}{A\arabic{table}}
\renewcommand{\thefigure}{A\arabic{figure}}

\section*{Limitations and Future Works}
In this work, we design a type of \textit{unnatural language} that is comprehensible to LLMs and demonstrate that it
contains useful features that could facilitate instruction tuning and even achieves on-par performance compared with the
corresponding natural language. Essentially, the unnatural language searching is a process that increases the entropy
of language (via replacing tokens in GCG) while trying to keep the semantic meaning of the original language. Therefore,
the result inevitably contains the tokens that appeared in the original natural string, which is not 100\% unnatural
ideally. We have also performed ablation study by eliminating the original tokens from the candidates and initialize the
searching control without natural tokens but found that the performance is significantly sub-optimal. We admit the
limitation of current searching methods leveraging GCG and conjecture that the unnaturalness and performance for
unnatural language could be further improved once there is more effective and efficient discrete space optimization
approach.

Besides, the efficiency for GCG-like searching methods is limited and it is expensive to search large
scale unnatural language version examples. This limits us from performing more comprehensive and large scale
experiments. Once the searching expense is mitigated, unnatural language could be further explored for practical usage.

In addition, we found that the unnatural language are not always generalizable across different tasks.
For example, fine-tuning on the unnatural version of GSM8K training set as in Sec.~\ref{sec:unnatural_patterns} does not
achieve on-par performance with the one fine-tuned on natural set in expectation. We suspect that this is because that GSM8K questions
are too complex (long) for GCG to search a compatible unnatural version. This also explains why the unnatural \SG in
Sec.~\ref{sec:ul_is_comprehensible_by_llms} only achieves 54\% accuracy in average across models.

\section{Algorithm Details}\label{sec:appen_alg}

Algorithm \ref{alg:searching_ul} presents the detailed implementation of the unnatural languages searching algorithm.

\begin{figure}[ht]
\vspace{-2mm}
	\scalebox{0.95}{
		\begin{minipage}{\linewidth}
			\begin{algorithm}[H]
				\renewcommand{\algorithmicrequire}{\textbf{Input:}}
				\renewcommand{\algorithmicensure}{\textbf{Output:}}
				\caption{Unnatural Languages Searching}\label{alg:searching_ul}
				\begin{algorithmic}[1]
					\REQUIRE Natural string \(S\), searching models \(\mathcal{M}\) and tasks \(\mathcal{T}\),
					batch size \(B\), \(k\), number of iterations $T$
					\STATE \texttt{// Initialize $x$ via shuffle words in $S$ and inject special characters.}
					\STATE{\textbf{Initialization: } \(x = \text{random\_inject}(\text{shuffle}(S))\)}
					\REPEAT
					\FOR{\(M \in \mathcal{M}\)}
					\STATE \texttt{// Tokenize $x$ through model $M$.}
					\STATE{\(\mathbf{x}_{1:n} = \text{Tokenize}_M(x)\)}
					\STATE \texttt{// Obtain top-k alternative tokens of each position in $\mathbf{x}_{1:n}$.}
					\STATE{$\mathbf{X}_{1:n}$ = $\text{Top-k}\big(\nabla_{\mathbf{x}_{1:n}}\sum_{t\in\mathcal{T}}\log P_M(S | \mathbf{x}_{1:n}, t)\big)$}
					\FOR{$b = 1,..., B$}
					\STATE \texttt{// Uniformly sample candidates.}
					\STATE{$\tilde{\mathbf{x}}^{(b)}_{1:n} = \mathbf{x}_{1:n}$}
					\STATE{$\tilde{\mathbf{x}}^{(b)}_{1:n}[i] = \text{Uni}(\mathbf{X}_{1:n}[i]),\;i = \text{Uni}([1:n])$}
					\STATE \texttt{// Decode tokens back string.}
					\STATE{$\tilde{x}^{(b)} = \text{Decode}_M(\tilde{\mathbf{x}}^{(b)}_{1:n})$}
					\ENDFOR
					\ENDFOR
					\STATE \texttt{// Select the best candidate.}
					\STATE{$\tilde{x}^{b^*} = \text{argmax}_b\sum_{M\in\mathcal{M}}\sum_{t \in
								\mathcal{T}}\log P_M(S | \tilde{x}^{(b)} , t)$}
					\STATE \texttt{// Replace the original string with the modified string.}
					\STATE{$x = \tilde{x}^{b^*}$}
					\UNTIL{Repeat for $T$ times}
					\ENSURE Equivalent unnatural string \(S'\)
				\end{algorithmic}
			\end{algorithm}
		\end{minipage}
	}
\end{figure}

\section{Algorithm Verificaiton Experiments}\label{sec:verification}

Here we perform a verification experiment to show that we searched unnatural strings could be translated back to natural
version. Specifically, we perform such translation task by appending a translation task description 'Translate the above
sentences into natural languages'. The translation performance is shown in Table~\ref{tab:translation_verification}. 
\emph{EM} denotes exact match; \emph{F1} denotes F1 score; and \emph{NLI} measures the semantic relationship between sentences (e.g., entailment, neutral, contradiction) using pre-trained models, offering a nuanced evaluation of meaning similarity. We treat 'entailment' as positive cases and compute the accuracy to obtain the final score.

\begin{table}[ht]
	\caption{Results of unnatural language to natural language translation task.}
	\centering
	\scalebox{0.95}{
		\begin{tabular}{l|ccc}
			\toprule
			\textbf{Dataset} & \textbf{EM} & \textbf{F1} & \textbf{NLI} \\\midrule
			\SC              & 0.5600      & 0.854       & 0.860        \\
			\SG              & 0.660       & 0.805       & 0.650        \\
			\bottomrule
		\end{tabular}}
	\label{tab:translation_verification}
\end{table}

\section{\SC and \SG Dataset Details}\label{sec:appen_prompt}

\subsection{\SC}

\textbf{Generation Details.} \;
We leverage GPT-3.5~\citep{john2023chatgpt} to generate context about non-existing entities
and their corresponding questions. The prompt is provided as follows in Table \ref{table:prompt}. To ensure the diversity
of the generated context, we generate \(1,000\) candidates in total and perform k-means
clustering according to the embeddings generated by a SOTA text embedding model to form 
100 clusters. Finally, we select 100 instances that are closest to each of the cluster centers. 

    



\begin{table}[ht]
	\caption{Prompt of \SC generation. }
	\centering
	\scalebox{0.9}{
		\begin{tabular}{p{18cm}}
			\toprule
			 Please generate 10 synthetic business or personal case for reading comprehension. The context information should be specific to a synthetic object, e.g., 'The company TechDouDou raised 1,000,000 fundings in Q4, 2023' instead of 'A company raised 1,000,000 fundings in Q4, 2023'. all data should be different from each other as much as possible. The case contains three parts: (1) context that provides specific information, where the length should be no longer than 40 characters; (2) a question that asked about that information; (3) the corresponding answer.
    
[Context]:The revenue of the company Countingstar for Q1 is 100,000\$.

[Question]: What is the revenue of Countingstar for Q1?

[Answer]: The revenue of Countingstar for Q1 is 100,000\$                                                                   \\
			\bottomrule
		\end{tabular}}
	\label{table:prompt}
\end{table}

\textbf{Dataset Details.} \; 
The dataset is generated by GPT-3.5. Each data contains a simple and synthetic context related to unexisted business or
personal as well as a couple of questions related to the context. Genearally, one question asks about the action of the
entity while the other question asks about the name of the enitity. For example, the context is \textit{"EcoGardens launched a new
	sustainable packing initiative"} and the two questions are \textit{"What recent initiative did EcoGradens launch?"} and
\textit{"Which company launched a new sustainable packing initiative?"}. To ensure the diversity of the generated
questions, we generated 1,000 contexts and leverage k-means to get 100 data points, each of which is closed to the centers
of a cluster. the cluster embeddings were generated by a SOTA embedding model in
\texttt{sentence-transformers}~\citep{reimers2019sentence}. For each data point, we manually create the correct answer
candidates for each question.

\subsection{\SG}~\label{sec:dataset_details}

\textbf{Dataset Details.} \;
The dataset serves as a more challenging dataset compared to \SC, where the data points are derived from
the test set of \texttt{GSM8K}. As a result, the context is more complex which often contains multiple entities and much
more information. Meanwhile, the answer of the question requires several steps of reasoning and the correct answer
could not be found in the context directly using simple keyword matching. Since our goal is to evaluate the ability of
unnatural language comprehension of LLMs, we do not want to introduce too complex QA paris which could not even be answered
correctly under natural language. To this end, we selected 100 questions from the original GSM8K test set considering the
context length and correctness of models.

\section{Unnatural QA Experiments Under Two-turn Dialogue Setting.}~\label{sec:appen_two}

To verify the model's
understanding of unnatural context and given that these are chat-based models,
we implement a dialogue format for context-based question-answering.
Specifically, each QA session consists of two turn. In the first
turn, we provide the context in unnatural language to the model, which responds
with ``OK, got it.'' In the second turn, we pose the question related to the
previously provided context and evaluate the model's response accuracy through
keyword exact matching. Detailed results are shown in Table \ref{tab:inference_on_ul_two_turn}. 
The results serves as a complimentary results of Table~\ref{tab:inference_on_ul}
and it further demonstrates that such unnatural languages is highly transferrable across models.

\begin{table*}[t]
	\centering
	\caption{Performance comparison for different contexts across different models on SynContextQA and SimGSM8K datasets
		under \emph{tow-turn dialogue setting}.
		All answers were generated under zero-shot setting without sampling. ``Direct'' refers to models used for unnatural language searching, while ``Transfer'' indicates the implementation of searched unnatural languages.}
	\scalebox{1.0}{
		\begin{tabular}{ll|cc>{\columncolor{lightblue}}c|cc>{\columncolor{lightblue}}c}
			\toprule
			                                                & \multirow{2}{*}{\textbf{\normalsize{Model}}} & \multicolumn{3}{c}{\textbf{SynContextQA}} \vline & \multicolumn{3}{c}{\textbf{SimGSM8K}}                                              \\\cmidrule(lr){3-8}
			                                                &                                              & Natural                                          & Shuf-InJ                              & Unnatural & Natural & Shuf-InJ & Unnatural \\
			\midrule
			\multirow{4}{*}{\textbf{\normalsize{Direct}}}   & Mistral-7B-Instruct                          & 0.92                                             & 0.47                                  & 0.92      & 0.85    & 0.20     & 0.42      \\
			                                                & Vicuna-7B                                    & 0.94                                             & 0.49                                  & 0.90      & 0.63    & 0.18     & 0.21      \\
			\cmidrule(lr){2-8}
			                                                & Average                                      & 0.93                                             & 0.48                                  & 0.91      & 0.74    & 0.19     & 0.32      \\
			\midrule
			\multirow{5}{*}{\textbf{\normalsize{Transfer}}} & Meta-Llama-3-8B-Instruct                     & 0.98                                             & 0.51                                  & 0.84      & 0.77    & 0.31     & 0.40      \\
			                                                & Gemma-2-9B-Instruct                          & 0.96                                             & 0.46                                  & 0.70      & 0.97    & 0.22     & 0.45      \\
			                                                & Meta-Llama-3-70B-Instruct                    & 0.98                                             & 0.70                                  & 0.92      & 1.00    & 0.41     & 0.73      \\
			                                                & GPT-3.5                                      & 0.98                                             & 0.68                                  & 0.92      & 0.92    & 0.38     & 0.49      \\
			                                                & GPT-4                                        & 0.99                                             & 0.64                                  & 0.91      & 0.96    & 0.32     & 0.48      \\
			\cmidrule(lr){2-8}
			                                                & Average                                      & 0.98                                             & 0.60                                  & 0.86      & 0.92    & 0.33     & 0.51      \\
			\bottomrule
		\end{tabular}}
	\label{tab:inference_on_ul_two_turn}
\end{table*}

\section{LLMs Understand the Dependency Structure of Unnatural Language.} \label{sec:appen_example}

Dependency parsing is one of the most commonly used techniques to analyze the syntactic structure of natural sentences.
Dependency parsing transfer a sentence into a tree, where each word/token is a node and the directed edges represent the
dependency, the end node (child) depends on, e.g. modifying or being arguments of, the source node (parent).
In general, a fake ROOT node is usually added to the whole tree such that every actual word have parents.
Dependency parsing serves as a fundamental technique for understanding the syntax of sentences. In this section, we are
curious about how LLMs understand the unnatural language. A more specific question is that how LLMs interpret the syntax
of unnatural language since they could understand it.

Recently, \citet{hewitt2019structural} showed that pretrained LMs', e.g. BERT~\citep{devlin2018bert}, output embeddings contains
the structural syntax information of the inputs. One can simply leverage the probing (a linear transformation)
techiques to extract such information and build a dependency syntax tree. Leveraging a modified version of the
so-called \textit{structural probing}, we train a linear head upon a freezed pre-trained model (e.g. \texttt{Llama-3-8B})
to predict the dependency syntax tree of the input sentence using natural language corpus as training source. Then we
leverage the trained head to predict the syntax tree of unnatural sentences. We introduce the structural probing method as follows.

A syntax tree of a sentence \(l\) is equal to a directed acyclic graph (DAG) \(G = (V, E)\), where \(V\) is the set of
words and \(E\) denotes the set of edges. to build the guidance for training, we compute a distance matrix \(\mathbf{d}
\in \mathbb{N}^{(|V|, |V|)}\) for the
graph. The element \(d_{ij}\) of the distance matrix is defined as the length of the shortest path between
node \(i\) and node \(j\). Particularly, if two nodes are adjacent, the distance is \(1\).
In the original paper, the authors omitted the root node for syntax tree when building the distance matrix, which
results in a symmetric matrix that loses the information of root node and edge directions. In our work, we
treat the root node as a concrete token (the BOS special token for any LLM tokenizer) and build the corresponding
distance matrix. Such distance matrix is an one-to-one mapping of the syntax tree since one can easily recover the
syntax tree since the root node is known and the tree is acyclic. In the following parts, we use
\(\mathbf{d}^l\in\mathbb{N}^{(|V_l|+1, |V_l|+1)}\) denoting the distance matrix for sentence \(l\).

Given a model's output embedding \(\mathbf{h}^l \in \mathbb{R}^(L, H)\) for sentence \(l\). We formalize the training
process as the following optimization problem:

\begin{equation}
	\min_{\mathbf{W}} \sum_l \frac{1}{|s^l|^2} \sum_{i,j}|\mathbf{d} - f_\mathbf{W}(\mathbf{h}_i^l, \mathbf{h}_j^l)|,
\end{equation}
where \(f_\mathbf{W} = (\mathbf{W}(\mathbf{h}_i^l-\mathbf{h}_j^l))^T(\mathbf{W}(\mathbf{h}_i^l-\mathbf{h}_j^l))\)
computes the squared eucliean distance between two nodes. \(|s^l|\) denotes the number of tokens in sentence
\(l\). In practice we train the probing head with the data from EN\_EWT training set of universal
dependency~\citep{silveira14gold}. 

\begin{figure}[!t]
	\begin{center}
		\resizebox{0.6\linewidth}{!}{
			\newcommand\blue[1]{\textcolor{RoyalBlue}{#1}}
\begin{forest}
	for tree={
	edge={thick, ->}, 
	s sep=1mm, 
	l sep=1mm 
	}
	[\textless{}\textbar{}begin\_of\_text\textbar{}\textgreater{} [\blue{proceeded}, edge={draw=RoyalBlue}[thus] [.
					[\blue{initiative}, edge={draw=RoyalBlue}[jekt] [\blue{launched}, edge={draw=RoyalBlue}[\blue{pack}, edge={draw=RoyalBlue}
											[um]] [\blue{One}, edge={draw=RoyalBlue}] [\blue{able}, edge={draw=RoyalBlue}]] [ierte [stoff] [deze] [Lng [y [...]] [ut] [abet]] [nouvelles]] [sentence
									[level [weiter [...]] [grammar [Parse]]] [former]]]] [\blue{G}, edge={draw=RoyalBlue}]
			[\blue{ens}, edge={draw=RoyalBlue} [\blue{Eco}, edge={draw=RoyalBlue}] [\blue{ard}, edge={draw=RoyalBlue}]]]]
\end{forest}}
	\end{center}
	\caption{A dependency syntax tree of unnatural sentence: "\textit{| EcoGardenslaz proceeded thus,- sust "able deix um
			nouvelles packstoff launchedierteutabetLng initiative. Onejekt y deze sentence former GETTRAN()->()\} grammar level:: EGB
			>>OK!. Parse Sie fast\{-itt weiter}", whose corresponding natural version is "\textit{EcoGardens launched a new
			sustainable packaging initiative}". The tree was generated by \texttt{Llama-3-8B} via structural
		probing~\citep{hewitt2019structural}. We annotate a sub-tree which reasonably represents the semantic meaning of
		"\textit{EcoGardens proceeded initiative that launched one sustainable pack.}"}\label{fig:dp_tree}
\end{figure}
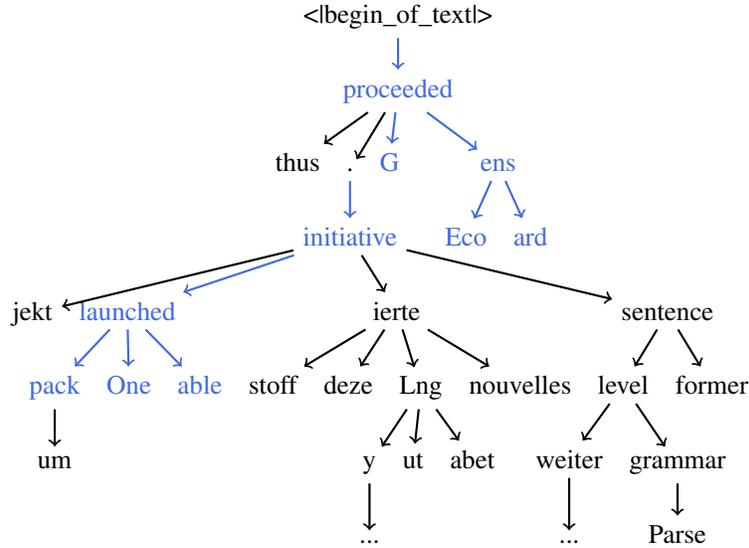

Then we use the trained probing head to predict the dependency syntax tree of unnatural sentences. We show an example in
Figure~\ref{fig:dp_tree}. As shown in Figure~\ref{fig:dp_tree}, the dependency tree clearly contains the syntax structure
whose semantic meaning is similar to the natural language. This indicates that LLMs could capture the syntax structure
of unnatural language that contains information akin to the natural language.

\section{More Experiments with additional baselines}

To further validate the effectiveness of the proposed unnatural languages, we conduct additional experiments comparing the dropping-token baseline with our method. Specifically, we employed saliency technique~\citep{mundhenk2020saliency} to retain the top percentage of the most influential tokens while discarding the rest—a common approach in Explainable AI for identifying which parts of the input most strongly impact a model’s prediction. The examples of baselines are shown in Table~\ref{tab:additional_baseline_examples} and the experiment results are presented in Table~\ref{tab:additional_baseline}.

We first compare the unnatural language outputs generated by the baseline and our approach. The baseline tends to preserve the word order and keywords of the original input, making it comparatively more human-readable. In contrast, our method generates outputs that are less comprehensible to humans while maintaining critical latent features important for LLMs.

Moreover, on the SimGSM8K dataset, our unnatural examples consistently lead to significantly higher performance across multiple models compared to the baseline. This demonstrates that our method results in examples that are more unnatural from a human perspective while still preserving the essential latent structure that LLMs rely on for reasoning.

\begin{table}[t]
\caption{Examples of baselines and unnatural language on SimGSM8K.}
\centering
\resizebox{\textwidth}{!}{
    \begin{tabular}{l l}
        \toprule
        \textbf{Method} & \textbf{Examples} \\
        \midrule
        pure\_top0.3 & Carly arms seastar \\
        pure\_top0.5 & Carlyfish5 arms each one seastar. \\
        pure\_top0.7 & Carly collected7 starfish5 arms each and one seastar arms. \\
        random\_injection\_top0.3 & \makecell[l]{Carly conclude Grudsignature)' nordMBERazed arms Python GorNEXT \\ seastar anime workshop Felixlights gardenearing} \\
        random\_injection\_top0.5 & \makecell[l]{Carly Yale embedding prospects Controlfish practicallyunsigned5 arms \\ each supp one seastarquote personnelscore AuthorsVal.} \\
        random\_injection\_top0.7 & \makecell[l]{Carly collected tra7 starfishNext bet5 arms each and one seastar\\ amplWM substantial hacer arms.} \\
        \midrule
        Unnatural (Ours) & \makecell[l]{\textbar{} Each and : algebra dinner! absolutely 7 do): shortly . seastar collectedthe\\ \textbackslash`' kW)\textdollar{}, one ! 5 ! 14\textbackslash` starfish with sic\}\}\_\textbackslash{}label Carly\} arms. Onehorailey\\ constructed WriteStatus(\textdollar{} \textdollar{}\textbackslash{}Toggle Zwezeichnung OK} \\
        Original Context & Carly collected 7 starfish with 5 arms each and one seastar with 14 arms. \\
        \bottomrule
    \end{tabular}
}
\label{tab:additional_baseline_examples}
\end{table}

\begin{table}[t]
\caption{Experiment results of different baselines on SimGSM8K. Pure refers to retaining only the most salient tokens. Random Injection (RI) denotes the addition of randomly selected tokens to increase the level of unnaturalness.}
\centering
\resizebox{\textwidth}{!}{
\begin{tabular}{lcccc}
\toprule
\textbf{Method} & \textbf{Mistral-7B-Instruct-v0.1} & \textbf{Meta-Llama-3-8B-Instruct} & \textbf{Gemma-2-9b-it} & \textbf{Meta-Llama-3-70B-Instruct} \\
\midrule
\textbf{pure\_top0.3} & 0.07 & 0.07 & 0.07 & 0.07 \\
\textbf{pure\_top0.5} & 0.10 & 0.06 & 0.12 & 0.14 \\
\textbf{pure\_top0.7} & 0.18 & 0.12 & 0.25 & 0.24 \\
\textbf{random\_injection\_top0.3} & 0.08 & 0.08 & 0.07 & 0.10 \\
\textbf{random\_injection\_top0.5} & 0.11 & 0.11 & 0.15 & 0.16 \\
\textbf{random\_injection\_top0.7} & 0.17 & 0.16 & 0.23 & 0.28 \\
\midrule
\textbf{Unnatural (Ours)} & \textbf{0.42} & \textbf{0.50} & \textbf{0.41} & \textbf{0.75} \\
\bottomrule
\end{tabular}
}
\label{tab:additional_baseline}
\end{table}




\end{document}